\RequirePackage[l2tabu,orthodox]{nag}

\documentclass{llncs} 

\PassOptionsToPackage{table,svgnames,dvipsnames}{xcolor}

\usepackage[utf8]{luainputenc}
\usepackage[T1]{fontenc}
\usepackage[sc]{mathpazo}
\usepackage[ngerman,english]{babel} 
\usepackage[autostyle]{csquotes}

\newcommand{\parencite}[1]{\cite{#1}}

\usepackage{amsmath}
\usepackage{graphicx}
\usepackage{scrhack} 
\usepackage{listings}
\usepackage{lstautogobble}
\usepackage{tikz}
\usepackage{pgfplots}
\usepackage{pgfplotstable} 
\usepackage{booktabs} 
\usepackage[final]{microtype}
\usepackage{caption}
\usepackage[hidelinks]{hyperref} 
\usepackage{ifthen} 
\usepackage{pdftexcmds} 
\usepackage{paralist} 
\usepackage{subfig} 
\usepackage{multirow} 
\usepackage{array} 
\usepackage{makecell} 
\usepackage{pdfpages} 
\usepackage{adjustbox} 
\usepackage{tablefootnote} 
\usepackage{threeparttable} 
\usepackage{pdflscape} 
\usepackage{mathtools}
\usepackage{bm}

\usepackage{amssymb}
\usepackage{pifont}
\usetikzlibrary{angles, calc, quotes, positioning, external, patterns, shapes}

\addto\extrasenglish{
}

\definecolor{TUMBlue}{HTML}{0065BD}
\definecolor{TUMSecondaryBlue}{HTML}{005293}
\definecolor{TUMSecondaryBlue2}{HTML}{003359}
\definecolor{TUMBlack}{HTML}{000000}
\definecolor{TUMWhite}{HTML}{FFFFFF}
\definecolor{TUMDarkGray}{HTML}{333333}
\definecolor{TUMGray}{HTML}{808080}
\definecolor{TUMLightGray}{HTML}{CCCCC6}
\definecolor{TUMAccentGray}{HTML}{DAD7CB}
\definecolor{TUMAccentOrange}{HTML}{E37222}
\definecolor{TUMAccentGreen}{HTML}{A2AD00}
\definecolor{TUMAccentLightBlue}{HTML}{98C6EA}
\definecolor{TUMAccentBlue}{HTML}{64A0C8}

\pgfplotsset{compat=1.16}
\pgfplotsset{
  cycle list={TUMBlue\\TUMAccentOrange\\TUMAccentGreen\\TUMSecondaryBlue2\\TUMDarkGray\\},
}

\graphicspath{
    {logos/}
    {figures/}
}

\renewcommand{\i}{\emph}
\renewcommand{\t}{\texttt}
\renewcommand{\b}{\textbf}

\newcommand{\dtControl}{\t{dtControl}}
\newcommand{\cruise}{\t{cruise}}

\newcolumntype{P}[1]{>{\centering\arraybackslash}p{#1}} 
\newcolumntype{M}[1]{>{\centering\arraybackslash}m{#1}} 
\newcolumntype{L}[1]{>{\raggedright\arraybackslash}m{#1}} 
\newcolumntype{R}[1]{>{\raggedleft\arraybackslash}m{#1}} 
\newtoggle{arxiv}
\newcommand{\arxivref}[2]{\iftoggle{arxiv}{#1}{#2}}
\toggletrue{arxiv}

\let\llncssubparagraph\subparagraph
\let\subparagraph\paragraph
\usepackage[compact]{titlesec}
\usepackage{titlesec}
\let\subparagraph\llncssubparagraph

\titlespacing*{\section}{0pt}{3ex plus 1ex minus 1ex}{2ex plus 0.5ex minus 0.5ex}
\titlespacing*{\subsection}{0pt}{2.5ex plus 0.5ex minus 1ex}{1ex plus 0.5ex minus 0.5ex}
\titlespacing*{\subsubsection}{0pt}{1ex plus 0.5ex minus 0ex}{1ex}

\begin{document}
	
	\title{Algebraically Explainable Controllers: Decision Trees and Support Vector Machines Join Forces}
	%
	%
	\author{Florian J\"ungermann \and
		Jan K\v{r}et\'{i}nsk\'{y}
		\and
	Maximilian Weininger\thanks{This research was funded in part by the German Research Foundation (DFG) projects 383882557 Statistical Unbounded Verification (SUV) and	427755713 Group-By Objectives in Probabilistic Verification (GOPro).}}
	
	%
	%
	\institute{Technical University of Munich, Germany}
	
	\pagestyle{plain}

\newcommand{\pics}{}

\maketitle

\begin{abstract}
	Recently, decision trees (DT) have been used as an explainable representation of controllers (a.k.a.~strategies, policies, schedulers). Although they are often very efficient and produce small and understandable controllers for discrete systems, complex continuous dynamics still pose a challenge. In particular, when the relationships between variables take more complex forms, such as polynomials, they cannot be obtained using the available DT learning procedures. In contrast, support vector machines provide a more powerful representation, capable of discovering many such relationships, but not in an explainable form. Therefore, we suggest to combine the two frameworks in order to obtain an understandable representation over richer, domain-relevant algebraic predicates. We demonstrate and evaluate the proposed method experimentally on established benchmarks.
	
	
	
\end{abstract}


\section{Introduction}\label{chapter:introduction}

Safe and efficient controllers for cyber-physical systems are hard to obtain manually, in particular in presence of both the discrete type of behaviour and continuous aspects such as complex dynamics in space and/or time. 
To that end, various model checking tools offer also an automatic \emph{controller synthesis} option, for instance \t{UPPAAL Stratego} \parencite{David2015}, \t{PRISM} \parencite{prism}, \t{SCOTS} \parencite{scots}, or \t{STORM} \parencite{storm}. 
In their most basic form, they use discretization to represent the continuous input space with a finite set of states. For each of those states, the synthesized controller describes which actions are allowed. So, the controller can be expressed explicitly as a lookup table, often with millions of rows.

There are two main issues with this representation. First, \emph{storing} such a large table can require several hundreds of megabytes of storage. However, the devices on which the controller should run are often embedded chips with very limited storage capacity. This makes it infeasible to store the entire lookup table on the device.
Second, the sheer \emph{size makes it impossible to understand} the behavior of the controller. The safety guarantees of the controller rely on the assumption that the formal model was correct and behaves as expected. To validate this and certify the quality, understanding the controller is crucial. For example, a non-permissive controller for the emergency braking system might try to immediately stop the car. This fulfils the safety requirement, as no crash can occur; however, it is not useful in a real application.
These flaws in the model can be detected if we can represent the safe controller in a succinct and explainable way.

\paragraph{Running Example}
To demonstrate our approach to face these issues, we have a closer look at the adaptive cruise control model (in short \cruise) from \cite{cruise}, which models a simplified emergency braking system for a car. Synthesizing a safe controller with \t{UPPAAL Stratego} gives us a file with more than six million lines, although previous work \cite{akmese} has shown that there is a way of formulating the safe behavior with a handful of sentences or equations. The goal of this paper is to find such a succinct and explainable representation automatically, utilizing techniques from machine learning.

\subsubsection{Controller Representation with Decision Trees}
Recently significant progress has been made \cite{counterexample,Ashok2019,sos,dtControl,dtcontrol2} with representing controllers using decision trees (a.k.a. nested if-then-else code) . Decision trees, e.g.~\cite{Mitchell1997}, are simple in structure, making them easy to understand, but still expressive enough to represent complex controllers. The open-source tool \dtControl\ \cite{dtcontrol2} takes advantage of that and offers an automated way of generating succinct decision trees. It can read controllers from many commonly used model checkers and implements various heuristics to minimize the size of the decision tree.

Traditionally, in a decision tree, the predicate in the decision node has the form $v_i \le c$ with some variable $v_i$ and some constant $c\in \mathbb{R}$. Such a split divides up the feature space with a hyperplane orthogonal to the feature axis $v_i$ thus giving it the name \i{axis-aligned} split \parencite{Mitchell1997}. Those splits are easy to understand and efficient to find, forming the basis of efficient decision-tree learning. 

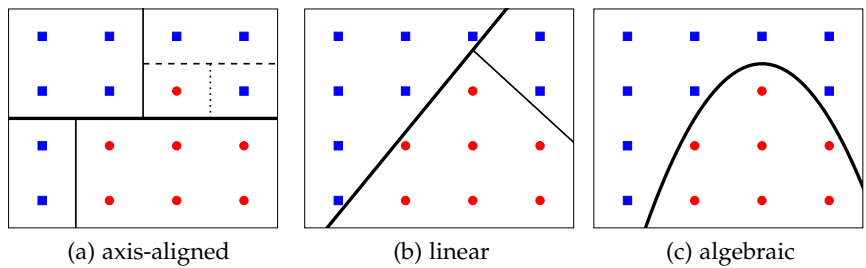
\begin{figure}
  \centering
  \subfloat[][axis-aligned]{
    \resizebox{0.3\textwidth}{!}{\begin{tikzpicture}
  \begin{axis}[
    ticks=none,
    xmin=0.5, xmax=4.5,
    ymin=0.5, ymax=4.5,
    width=0.5\textwidth ,
	  scatter/classes={
      a={mark=square*,blue},
      b={mark=*,red}
    }
  ]
    \addplot[
      scatter,
      only marks,
      scatter src=explicit symbolic
    ]
    table[meta=label]
    {	x   y   label
    	1   1   a 
    	1   2   a
    	1   3   a
    	1   4   a
    	2   1   b
    	2   2   b
    	2   3   a
    	2   4   a
    	3   1   b
    	3   2   b
    	3   3   b
    	3   4   a
    	4   1   b
    	4   2   b
    	4   3   a
    	4   4   a
    };
    
    \draw[ultra thick] (0, 2.5) -- (5, 2.5);
    \draw[thick] (1.5, 0) -- (1.5, 2.5);
    \draw[thick] (2.5, 2.5) -- (2.5, 5);
    \draw[dashed, thick] (2.5, 3.5) -- (5, 3.5);
    \draw[dotted, thick] (3.5, 2.5) -- (3.5, 3.5);
  \end{axis}
\end{tikzpicture}}
    \label{fig:predicate_compare_aa}
  }
  \subfloat[][linear]{
    \resizebox{0.3\textwidth}{!}{\begin{tikzpicture}
  \begin{axis}[
    ticks=none,
    xmin=0.5, xmax=4.5,
    ymin=0.5, ymax=4.5,
    width=0.5\textwidth ,
	  scatter/classes={
      a={mark=square*,blue},
      b={mark=*,red}
    }
  ]
    \addplot[
      scatter,
      only marks,
      scatter src=explicit symbolic
    ]
    table[meta=label]
    {
    x   y   label
    1   1   a 
    1   2   a
    1   3   a
    1   4   a
    2   1   b
    2   2   b
    2   3   a
    2   4   a
    3   1   b
    3   2   b
    3   3   b
    3   4   a
    4   1   b
    4   2   b
    4   3   a
    4   4   a
	};
    \draw[ultra thick] (0.5, 0) -- (4, 5.25);
    \draw[thick] (3, 3.75) -- (5, 1.5);
  \end{axis}
\end{tikzpicture}}
    \label{fig:predicate_compare_linear}
  }
  \subfloat[][algebraic]{
    \resizebox{0.3\textwidth}{!}{\begin{tikzpicture}
  \begin{axis}[
    ticks=none,
    xmin=0.5, xmax=4.5,
    ymin=0.5, ymax=4.5,
    width=0.5\textwidth ,
	  scatter/classes={
      a={mark=square*,blue},
      b={mark=*,red}
    }
  ]
    \addplot[
      scatter,
      only marks,
      scatter src=explicit symbolic
    ]
    table[meta=label]
    {
    	x   y   label
    	1   1   a 
    	1   2   a
    	1   3   a
    	1   4   a
    	2   1   b
    	2   2   b
    	2   3   a
    	2   4   a
    	3   1   b
    	3   2   b
    	3   3   b
    	3   4   a
    	4   1   b
    	4   2   b
    	4   3   a
    	4   4   a
    };
    \addplot [
      domain=1:5,
      restrict y to domain=0:5,
      samples=50,
      ultra thick,
    ] {-(x-3)^2 + 3.5};
  \end{axis}
\end{tikzpicture}}
    \label{fig:predicate_compare_algebraic}
  }
  \caption{Example showing how different types of predicates can separate a dataset.}
  \label{fig:predicate_compare}
\end{figure}

However, axis-aligned predicates are incapable of capturing more complex relationships as seen in \autoref{fig:predicate_compare_aa}. In this toy example, 5 predicates are needed to separate the red from the blue labels. For a real-world dataset with thousands of data points, this behavior can be even more extreme. For that reason, \dtControl\ also supports \i{linear predicates} as proposed by \parencite{Murthy1994}. These splits are still hyperplanes but now can have arbitrary orientations (see \autoref{fig:predicate_compare_linear}). This makes the predicates harder to find, more difficult to comprehend as more variables are involved, but can ultimately give significantly smaller decision trees.

Extending the notion of using more complex decision predicates, the newest version of \dtControl\ allows the use of \i{algebraic predicates} \parencite{weinhuber,dtcontrol2} (see \autoref{fig:predicate_compare_algebraic}). A domain expert can provide arbitrary closed-form mathematical expressions that are then used in the decision tree construction. It is even possible to leave some constants unspecified and \dtControl\ will find suitable values for those.

\paragraph{Limitations}
While \dtControl\ has already greatly reduced the size of the controllers in many benchmarks, it still has room for improvement. Most of the implemented heuristics for continuous systems rely on clever ways of determinizing a non-deterministic controller. This means we start with a (possibly most) permissive controller, i.e.\ a controller that can permit several safe actions per state. Then for every state we dynamically select one action, making the choice in each state deterministic. This allows to represent the resulting strategy more succinctly.
However, in some instances such as the \cruise\ model, we want to keep the permissiveness of the controller, for example to give a human driver the maximum amount of freedom.

For accurately representing the most-permissive controller for the \cruise\ model without any determinization heuristics, \dtControl\ needs several hundred decision nodes and the resulting decision tree is hardly explainable. When providing the right domain knowledge, significantly smaller decision trees can be found by using algebraic predicates \parencite{akmese,weinhuber}. However, so far, the supplied domain knowledge had to be tailored to the problem by hand.

\subsubsection{Our Contributions}
We addresses these limitations by proposing, demonstrating, and critically evaluating two approaches:
\begin{itemize}
  \item First, we explore how to automate the generation of relevant algebraic predicates for decision trees (DT) using \emph{domain knowledge}.
  \item Second, we use support vector machines (SVM) to learn the predicates directly from \emph{data}, resulting in more accurate but less understandable predicates. Subsequently, we show how to improve their explainability.
  \item We experimentally evaluate this combination of DT and SVM learning on 28 case studies and analyze the results. In particular, we receive an explainable DT with only 5 decision nodes for the \cruise\ example.
\end{itemize}

\subsubsection{Related Work}\label{chapter:related_work}
This work extends the open-source tool \dtControl\ that was first presented in \parencite{dtControl}, covered in detail in \parencite{jackermeier}, and since then has been extended significantly \parencite{dtcontrol2}.
Adapting techniques from \i{machine learning} and \i{formal verification},
we combine the insights we receive from the controller data with the domain knowledge to construct smaller and more explainable decision trees.

There have been several approaches using non-linear predicates in decision trees. \parencite{Ittner1996} explicitly constructs new features by combining existing ones (for example take their product or ratio) while \parencite{Bennett1998} explicitly uses SVMs inside the decision nodes. Both sources focus on the decision tree's ability to generalize but not on the explainability. Specifically, they do not explicitly reconstruct algebraic predicates from the SVM.

In previous versions of \dtControl\ \parencite{weinhuber,dtcontrol2}, \i{curve fitting} \parencite{curvefittingBook} has been used to find undetermined coefficients in algebraic splitting predicates. This approach is based on regression analysis and uses \i{least square fitting} \parencite{Levenberg1944,Marquardt1963}. In our work, however, we use the predicates to separate the data rather than fitting it. For a more detailed comparison, see \arxivref{\autoref{sec:problem_with_curve_fitting}}{\textcolor{red}{TODO}}.


Typically, \i{binary decision diagrams} (BDDs) \cite{bdd} are used to represent controllers in a compressed way \cite{scots,Zapreev2018}. As BDDs can only represent a binary function $\{0,1\}^n \rightarrow \{0, 1\}$, this approach requires us to encode the list of state-action pairs of the controller in binary variables. As a result, the BDDs are hardly explainable. Additionally, the size of the BDD heavily depends on the variable ordering. Finding an optimal ordering is NP-complete \cite{Bollig1996} and currently known heuristics struggle with high-dimensional inputs.

\i{Algebraic decision diagrams} \cite{ADD} extend BDDs to support representing a function $\{0, 1\}^n \rightarrow S$ with $S \subset \mathbb{N}$. They have been used for representing controllers in, e.g., \cite{APRICODD}. However, they suffer from the same issues we discussed for BDDs.

\section{Preliminaries}\label{chapter:preliminaries}

The paper is concerned with representing controllers, which we thus now define.
While the specific type of dynamics of the system is irrelevant, we assume the states of the system are given by values of state variables:

\begin{definition}[Controller]
	For a model $\mathcal{M}$ with states $\mathcal{S}$ and actions $\mathcal{A}$, a \emph{controller} $C : \mathcal{S} \rightarrow 2^{\mathcal{A}}$ selects for every state $s\in \mathcal{S}$ a set of (so-called \emph{safe}) actions $C(s) \subseteq \mathcal{A}$.
   Moreover, we assume the set of states is $\mathcal{S} \subseteq \prod_{i=1}^M \mathrm{Dom}(v_i)$, where, for $1\le i\le M \in \mathbb{N}$, $v_i$ is a \emph{state variable} with domain $\mathrm{Dom}(v_i)$ and $\prod$ denotes the cartesian product.
\end{definition}

Note that this definition allows for \i{permissive} controllers that can provide multiple possible actions for a state. 
Further, we call a controller \i{deterministic} when $|C(s)| = 1$ for all $s\in \mathcal{S}$, meaning it chooses exactly one possible action in every state.%
\footnote{ 
Also note that according to our definition, the controller's decision is solely based on its current state, not its past states. In practice, this limitation can often be circumvented by encoding additional information about the past into the state. For example, the decision of whether to water the plants may depend on the precipitation of the last three days. Then we can model our current state as a tuple $(p_1, p_2, p_3)$ where $p_i$ describes the precipitation $i$ days ago.
}

\subsection{Representing Controllers by Decision Trees.}

\begin{definition}[Decision Trees]
  A \i{decision tree} $T$ is defined as follows:
  \begin{itemize}
    \item $T$ is a rooted full binary tree, meaning every node either is an \i{inner node} and has exactly two children, or is a \i{leaf node} and has no children.
    \item Every inner node $v$ is associated with a decision predicates $\alpha_v$. A decision predicate (or just predicate) is a boolean function $S \rightarrow \{0, 1\}$ over the input $S$.
    \item Every leaf $\ell$ is associated with an output label $a_\ell \in A$.
  \end{itemize}
\end{definition}

For learning a decision tree, numerous methods exists such as \t{CART} \cite{cart}, \t{ID3} \cite{id3}, and \t{C4.5} \cite{c45}. In principle, they all evaluate different \emph{predicates} (see \autoref{sec:predicates}) by calculating some impurity measure (see \autoref{sec:impurityMeasure}) and then greedily pick the most promising one before splitting the dataset on that predicate and recursively continuing with the two children.

A decision tree represents a function as follows: every input vector $\vec{x} \in S$ is evaluated by starting at the root of $T$ and traversing the tree until we reach a leaf node $\ell$. Then, the label of the leaf $a_\ell$ is our prediction for the input $\vec{x}$. When traversing the tree, at each inner node $v$ we decide at which child to continue by evaluating the decision predicate $\alpha_v$ with $\vec{x}$. If the predicate evaluates to true, we pick the left child, otherwise, we pick the right one. 

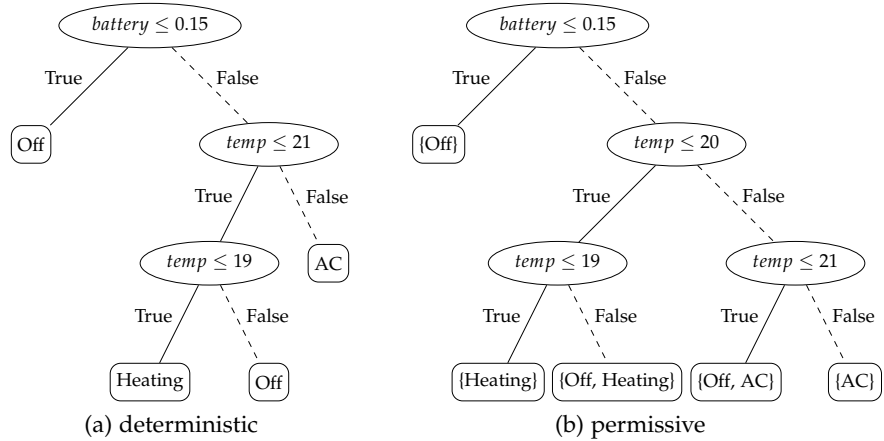
\begin{figure}[ht]
	\centering
	\subfloat[][deterministic]{
		\label{fig:dt_example_deterministic}
		\resizebox{!}{5.3cm}{\tikzstyle{dn} = [draw, ellipse]
\tikzstyle{leaf} = [draw, rounded corners=5pt, minimum height=2em]
\tikzstyle{t} = []
\tikzstyle{f} = [dashed]
\newcommand{\writesub}[1]{\\\small{$\boldsymbol{#1}$}}
\begin{tikzpicture}
  \node[dn] at (2, 0)   (root)  {$battery \le 0.15$};
  \node[dn] at (4, -2) (r)     {$temp \le 21$};
  \node[dn] at (3, -4)  (rl)    {$temp \le 19$};
  
  \node[leaf] at (0, -2)  (l)     {Off};
  \node[leaf] at (5, -4) (rr)    {AC};
  \node[leaf] at (2, -6)  (rll)   {Heating};
  \node[leaf] at (4, -6)  (rlr)  {Off};

  \draw[t] (root) -- node[anchor=east, yshift=4pt]{True}  (l);
  \draw[f] (root) -- node[anchor=west, yshift=4pt]{False} (r);
  \draw[t] (r)    -- node[anchor=east, yshift=4pt]{True}  (rl);
  \draw[f] (r)    -- node[anchor=west, yshift=4pt]{False} (rr);
  \draw[t] (rl)   -- node[anchor=east, yshift=4pt]{True}  (rll);
  \draw[f] (rl)   -- node[anchor=west, yshift=4pt]{False} (rlr);

\end{tikzpicture}}
	}
	\subfloat[][permissive]{
		\label{fig:dt_example_permissive}
		\resizebox{!}{5.3cm}{\tikzstyle{dn} = [draw, ellipse]
\tikzstyle{leaf} = [draw, rounded corners=5pt, minimum height=2em]
\tikzstyle{t} = []
\tikzstyle{f} = [dashed]
\tikzstyle{nf} = [anchor=west, yshift=4pt]
\tikzstyle{nt} = [anchor=east, yshift=4pt]
\newcommand{\writesub}[1]{\\\small{$\boldsymbol{#1}$}}
\begin{tikzpicture}
  \node[dn] at (2, 0)   (root)  {$battery \le 0.15$};
  \node[dn] at (4, -2) (r)     {$temp \le 20$};
  \node[dn] at (2, -4)  (rl)    {$temp \le 19$};
  \node[dn] at (6, -4) (rr)    {$temp \le 21$};
  
  \node[leaf] at (0, -2)  (l)     {\{Off\}};
  \node[leaf] at (1, -6)  (rll)   {\{Heating\}};
  \node[leaf] at (3, -6)  (rlr)   {\{Off, Heating\}};
  \node[leaf] at (5, -6) (rrl)   {\{Off, AC\}};
  \node[leaf] at (7, -6) (rrr)   {\{AC\}};

  \draw[t] (root) -- node[nt]{True}  (l);
  \draw[f] (root) -- node[nf]{False} (r);
  \draw[t] (r)    -- node[nt]{True}  (rl);
  \draw[f] (r)    -- node[nf]{False} (rr);
  \draw[t] (rl)   -- node[nt]{True}  (rll);
  \draw[f] (rl)   -- node[nf]{False} (rlr);
  \draw[t] (rr)   -- node[nt]{True}  (rrl);
  \draw[f] (rr)   -- node[nf]{False} (rrr);

\end{tikzpicture}}
	}
	\caption{An example of how a decision tree can represent a controller. (a) shows a determinized controller, (b) a permissive one with multiple safe actions at some states.}
	\label{fig:dt_example}
\end{figure}

When we represent a controller with a decision tree, our input data is the set of states and an output labels describe a subset of safe actions.
\autoref{fig:dt_example} shows a decision-tree representation of a deterministic and a permissive controller of a battery-powered temperature control system.

\subsection{Impurity Measures} \label{sec:impurityMeasure}
To evaluate how promising a predicate $\alpha$ is, \dtControl\ implements different impurity measures.
The most classic impurity measure is \i{entropy}, originating from information theory. It measures how much uncertainty is left in a dataset. If the dataset is dominated by one specific label, the entropy is low, whereas a heterogeneous dataset has a higher entropy. For a dataset $X$ with $N$ data points and the label set  $B$, let $n(X, y)$ be the number of data points in $X$ with the label $y \in B$. Then the entropy $H(X)$ is defined as:
\begin{align}
  H(X) = - \sum_{y\in B} \frac{n(X, y)}{N} \log_2\left(\frac{n(X, y)}{N}\right)
\end{align}
To evaluate a predicate $\alpha$, we calculate the remaining entropy after the split. If $\alpha$ is a binary split and partitions $X$ into $X = X_l \uplus X_r$, we define
\begin{align}
  H(\alpha, X) = \frac{|X_l|}{|X|} H(X_l) + \frac{|X_r|}{|X|} H(X_r)
\end{align}

\subsection{Predicates in Decision Trees}\label{sec:predicates}
In every decision node of our tree, a predicate function $A \rightarrow \{0,1\}$ is used to decide at which child we continue. We distinguish three categories of predicates according to their complexity:

\paragraph{Axis-Aligned Predicates}
are the simplest and by far the most commonly used type of predicates. They have the form $v_i \le c$ for a constant $c\in\mathbb{R}$. Geometrically speaking, the function $v_i = c$ describes a hyperplane orthogonal to the $v_i$ axis, intersecting at $v_i = c$. That is why they are called \i{axis-aligned} predicates \parencite{Mitchell1997}.

For finding the best axis-aligned predicate, we make the simple observation that for a feature $v_i$ with $k$ different values, there are only $k-1$ different relevant values for $c$. So we can simply evaluate all possible predicates for all features $v_i$ and select the most promising one.

\paragraph{Linear Predicates}
\parencite{Murthy1994} or sometimes called \i{oblique predicates} have the form $\sum_i a_iv_i \le c$ with $a_i, c \in \mathbb{R}$. This linear combination of different features describes a hyperplane with arbitrary orientation. Hence they are more expressive but it also makes it harder to find optimal predicates. Algorithms used to find suitable predicates include the OC1 algorithm \cite{Murthy1994}, logistic regression \cite[Chapter 4.4]{Hastie2009}, and support vector machines (SVM) \cite[Chapter 12]{Hastie2009} -- a machine-learning technique using a hyperplane to split a dataset into two partitions.
A hyperplane can be formally defined by the orthogonal vector that goes through the origin $\vec{w}$ and the distance from the origin $b$. Then it can be used as a classifier in the following way:
\begin{align}
	c: \mathbb{R}^M &\rightarrow \{-1, 0, 1\}\nonumber\\
	\vec{x} &\mapsto \text{sgn}(\vec{w}\cdot \vec{x} - b) \label{eq:decisionFunction}
\end{align}
where $\text{sgn}$ is the sign function.

\paragraph{Algebraic Predicates}
as outlined in \cite{akmese} and implemented in \cite{weinhuber,dtcontrol2}, are even more powerful predicates, which can reduce the size of the decision tree and improve explainability. \i{Algebraic predicates} allow any closed-form expressions and hence they are the most expressive. For the same reason, automatically generating good algebraic splitting predicates is difficult and typically requires human guidance.

\paragraph{}
\autoref{fig:predicate_compare} shows how the three types of predicates work on a toy dataset. As expected, the more expressive the predicate is, the fewer predicates are needed to build a perfect classifier.

\section{Running Example}\label{chapter:motivating_example}
In the cruise control model of \cite{cruise}, we want to control a car so that it will not crash into another vehicle in front. As a secondary objective, the car should drive as fast as possible, thereby minimizing the distance between both cars.

\begin{figure}[ht]
  \centering
  \includegraphics[scale=0.8]{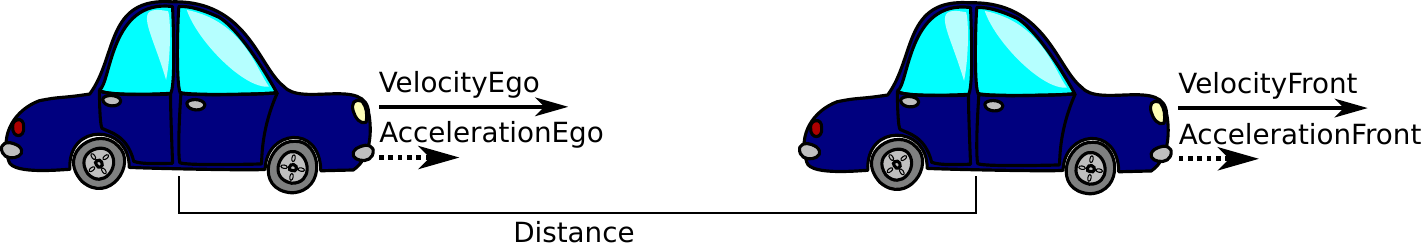}
  \caption{An illustration of the cruise model. Source: \cite{cruise}}
  \label{fig:cruise_illustration}
\end{figure}

The model is illustrated in \autoref{fig:cruise_illustration}. We only consider two vehicles, our vehicle called \i{ego} and the next vehicle in front of us called \i{front}. We drive on a single lane without cars entering or leaving, therefore this constellation does not change. The state of the system is modelled by the velocities $v_e$, $v_f$ of the cars and their relative distance $d_r$; the safety criteria $d_r \ge d_{safe}$ should hold at every state. In the model, both cars choose a constant acceleration $a_e$, $a_f$ for the duration of one time step $t_1$. Then, the new state ($d_r', v_e', v_f'$) is given by
\begin{align}
  v_e' &= v_e + a_e t_1\\
  v_f' &= v_f + a_f t_1\\
  d_r' &= d_r + (v_f-v_e) t_1 + \frac{1}{2}(a_f - a_e) t_1^2
\end{align}
The model restricts the domains of the accelerations to $a_e, a_f \in \{-2, 0, 2\}$ describing the three actions deceleration, neutral, and acceleration. Similarly, the cars have a bounded minimum and maximum velocity $v_{min}, v_{max}$ and the distance sensor has a limited reach of $d_{max}$. Depending on the values of these parameters, the size of the generated controller changes considerably.
For technical details, see \arxivref{Appendix~\ref{app:runningExample}}{\cite[Appendix A]{techreport}}.

\subsubsection{Insufficiency of the Current Solutions}
To illustrate the issues with the current solutions, we consider the dataset \t{cruise\_250} (see \arxivref{Appendix \ref{sec:cruiseParameters}}{\cite[Appendix A.2]{techreport}}). Here, the model checker \t{UPPAAL Stratego} \cite{David2015} generates a controller file with over 400MiB comprising 320,523 states and 961,569 state-action pairs. Representing it with a binary decision diagram \cite{bdd} still uses over 1,800 nodes. With \dtControl\ and axis-aligned or linear splits, we can get a decision tree with 869 or 369 nodes respectively which is still far too large to be understandable. Using the determinization heuristics discussed in \cite{dtControl}, we find a decision tree with only 3 nodes. Unfortunately, this determinized controller is of little use -- it simply lets the car decelerate until it reaches minimal velocity. Of course, this behavior satisfies the safety criteria but is not helpful in the real world.
To also fulfill the secondary objective of minimizing the relative distance we have two options. We can pre-determinize the controller by always picking the largest safe acceleration or we keep the maximal permissiveness. In the latter case, the cruise controller acts as an emergency braking system by letting the human driver choose any action as long as it is a safe one.

\subsubsection{Handwritten Strategy}
As shown in \cite{akmese}, there is a decision tree representing the most-permissive controller for the \cruise\ example with just 11 nodes. Yet, there has not been a way of automatically generating it with \dtControl\ so far.
The natural predicate for the decision to be made can be derived from the basic kinematics  of the model, and hence turns out to be the quadratic polynomial given in \autoref{eq:handcraftedPoly}. 
See \arxivref{Appendix~\ref{app:hand}}{\cite[Appendix A.3]{techreport}} for the derivation.

{
	\begin{align} \label{eq:handcraftedPoly}
		\frac{1}{2a_{min}} v_e^2 
		&- \frac{1}{2a_{min}} v_f^2 
		- \left(\frac{v_{min}}{a_{min}} + t_1(1 - \frac{a_{max}}{a_{min}})\right) v_e 
		+ \frac{v_{min}}{a_{min}} v_f 
		-\notag \\
		& \left(1 - \frac{a_{max}}{a_{min}}\right) t_1(t_1 a_{max} - v_{min}) 
		+\, d_r
		~~~~\ge ~~~~ d_{safe}
	\end{align}
}
\section{Predicates From Domain Knowledge}\label{chapter:pred_domain_knowledge}
As discussed in \cite{akmese,dtcontrol2}, we depend on domain knowledge provided by human experts to generate helpful splitting predicates. Providing very specific equations like the handcrafted predicate in \autoref{eq:handcraftedPoly} is inherently tedious and error prone. 
Thus, we aim to be able to synthesize the specific predicates from general domain knowledge, in our example the velocity and distance relations:
\begin{subequations}\label{eq:domainKnowledge}
\begin{align}
  v &= at\\
  d &= \frac{1}{2} a t^2 + vt
\end{align}
\end{subequations}

We proceed as follows.
Let $P$ be the set of physical quantities appearing in the general domain knowledge equations. 
In our concrete case, we have $P = \{d, v, a, t\}$, where these letters denote \textbf{d}istance, \textbf{v}elocity, \textbf{a}cceleration, and \textbf{t}ime. Then, let $S = \{ v_e, v_f, d_r\}$ be the set of state variables and $ C = \{d_{safe}, v_{min},\allowbreak v_{max},\allowbreak a_{acc},\allowbreak a_{neu},\allowbreak a_{dec}, t_1\}$ be a set of constants describing the minimum safety distance, the minimum and maximum velocities of the cars, the acceleration values corresponding to the actions \enquote{accelerate}, \enquote{neutral}, \enquote{decelerate}, and the duration of one time step. We observe that every constant and state variable describes exactly one physical quantity. We define the function $\rho_p (X)$ that returns the subset of entities of the set $X$ that are associated with the physical quantity $p\in P$. For example, $\rho_{d}(S \cup C) = \{d_r, d_{safe}\}$. Our approach can be described as following:
\begin{enumerate}
  \item Initialize $V_p \mathrel{:=} \rho_p(S\cup C)$ for all $p\in P$. Each $V_p$ contains the relevant values for the physical quantity $p$.
  \item For every equation from the domain knowledge (\ref{eq:domainKnowledge}), solve it for every physical quantity. This gives a set of equations in the form $p = f(P\setminus \{p\})$ for $p\in P$ that we call \i{base identities}. For example, the base identities for the acceleration are: $a = \frac{v}{t}$ and $a = \frac{2(d-vt)}{t^2}$. A complete list of all 8 base identities is in \arxivref{Appendix \ref{sec:allBaseIdentities}}{\cite[Appendix B.1]{techreport}}.
  \item For every physical quantity $p$ and every pair of values $x_1, x_2 \in V_p$, add $x_1 + x_2$ and $x_1 - x_2$ to $V_p$.
  \item For every base identity $\alpha$ associated with the quantity $p_\alpha$, and for every possible substitution function $\sigma$ that maps physical quantities $p\in P$ to values $x \in V_p$, add $\sigma(\alpha)$ to $V_{p_\alpha}$.
\end{enumerate}
Steps 3 and 4 can be repeated, thereby creating increasingly complex expressions. As an example, let us see how we can generate the value $d_{one}$ describing the difference in distance after one time step if the ego vehicle accelerates and the front vehicle decelerates. In Step 3 we add $a_{min} - a_{max}$ to $V_a$, as well as $v_f - v_e$ to $V_v$. In Step 4 we use the base identity $\alpha: d = \frac{1}{2}at^2 + vt$ with the substitutions
\begin{align*}
  \sigma(a) &= a_{min} - a_{max}\\
  \sigma(t) &= t_1\\
  \sigma(v) &= v_f - v_e
\end{align*}
and we receive the expression for $d_{one}$: $$d_{one} = \frac{1}{2}(a_{min} - a_{max})t_1^2 +(v_f - v_e)t_1$$

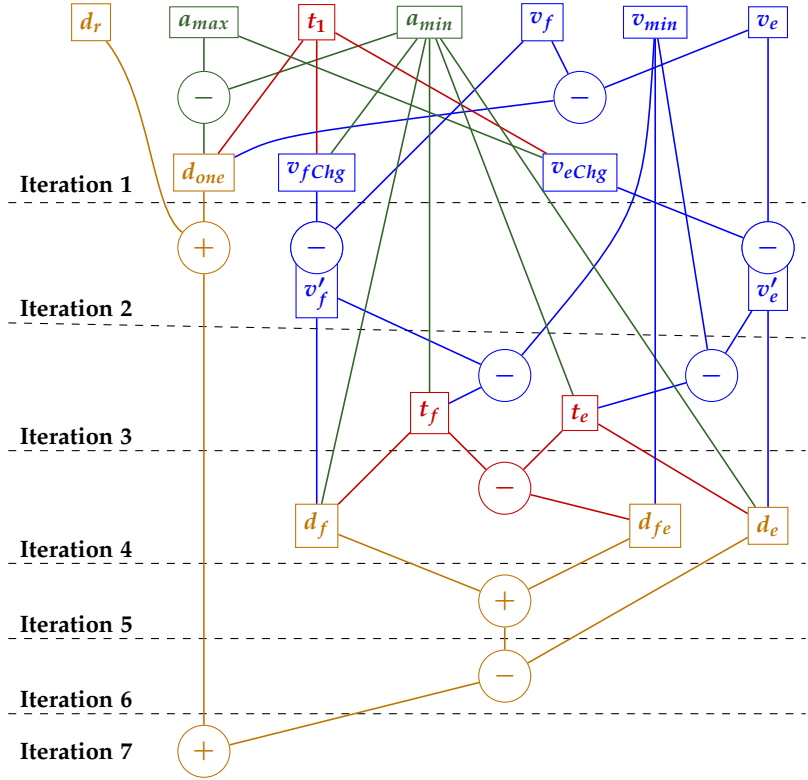
\begin{figure}[th]
	\centering
	\definecolor{mygreen}{HTML}{33652b}
\definecolor{myorange}{HTML}{c07400}
\definecolor{myred}{HTML}{c80000}
\tikzstyle{sum} = [circle, draw, fill=white]
\tikzstyle{term} = [draw, fill=white]
\tikzstyle{termV} = [term, draw=blue, text=blue]
\tikzstyle{termA} = [term, draw=mygreen, text=mygreen]
\tikzstyle{termS} = [term, draw=myorange, text=myorange]
\tikzstyle{termT} = [term, draw=myred, text=myred]
\tikzstyle{arr} = [semithick]
\tikzstyle{arrV} = [arr, blue]
\tikzstyle{arrA} = [arr, mygreen]
\tikzstyle{arrS} = [arr, myorange]
\tikzstyle{arrT} = [arr, myred]
\tikzstyle{subminbox} = [draw, rectangle, align=center, fill=white]
\newcommand{\writesub}[1]{\\\small{$\boldsymbol{#1}$}}
\begin{tikzpicture}
  \node[termS] at (-1.5, 10) (dr)   {$\boldsymbol{d_r}$};
  \node[termA] at (   0, 10) (aMax) {$\boldsymbol{a_{max}}$};
  \node[termT] at ( 1.5, 10) (t1)   {$\boldsymbol{t_1}$};
  \node[termA] at (   3, 10) (aMin) {$\boldsymbol{a_{min}}$};
  \node[termV] at ( 4.5, 10) (vF)   {$\boldsymbol{v_f}$};
  \node[termV] at (   6, 10) (vMin) {$\boldsymbol{v_{min}}$};
  \node[termV] at ( 7.5, 10) (vE)   {$\boldsymbol{v_e}$};

  \node[sum, termA] at (  0, 9) (aRel) {$\boldsymbol{-}$};
  \node[sum, termV] at (  5, 9) (vRel) {$\boldsymbol{-}$};

  \node[termS] at (   0, 8) (sOne) {$\boldsymbol{d_{one}}$};
  \node[termV] at (1.5, 8) (vfChg) {$\boldsymbol{v_{fChg}}$};
  \node[termV] at (   5, 8) (veChg) {$\boldsymbol{v_{eChg}}$};

  \draw[arrA] (aMax) -- (aRel);
  \draw[arrA] (aMin) -- (aRel);
  \draw[arrV] (vF)   -- (vRel);
  \draw[arrV] (vE)   -- (vRel);

  \draw[arrA] (aRel) -- (sOne);
  \draw[arrT] (t1)   -- (sOne);
  \draw[arrV] (vRel) ..controls(1, 8.5).. (sOne);

  \draw[arrA] (aMin) -- (vfChg);
  \draw[arrT] (t1)   -- (vfChg);
  \draw[arrA] (aMax) -- (veChg);
  \draw[arrT] (t1)   -- (veChg);

  \node at (-1.7, 7.85) {\small{\textbf{Iteration 1}}};
  \draw[dashed] (-2.6, 7.6) -- (8, 7.6);

  \node[sum, termS] at (  0, 7) (dSum) {$\boldsymbol{+}$};
  \node[sum, termV] at (1.5, 7) (vFnew) {$\boldsymbol{-}$};
  \node[sum, termV] at (7.5, 7) (vEnew) {$\boldsymbol{-}$};
  \node[subminbox, termV] [below = -0.65em of vFnew] (vFnewS)
    {\writesub{v_f'}};
  \node[subminbox, termV] [below = -0.65em of vEnew] (vEnewS)
    {\writesub{v_e'}};
  \node[sum, termV] at (1.5, 7) (vFnew) {$\boldsymbol{-}$};
  \node[sum, termV] at (7.5, 7) (vEnew) {$\boldsymbol{-}$};

  \draw[arrS] (sOne) -- (dSum);
  \draw[arrS] (dr) ..controls (-0.7, 9) and (-0.7, 7.5).. (dSum);
  \draw[arrV] (vfChg) -- (vFnew);
  \draw[arrV] (vF)    -- (vFnew);
  \draw[arrV] (veChg) -- (vEnew);
  \draw[arrV] (vE)    -- (vEnew);

  \node at (-1.7, 6.2) {\small{\textbf{Iteration 2}}};
  \draw[dashed] (-2.6, 6) -- (8, 5.8);


  \node[sum, termV] at (4, 5.3) (vFRel) {$\boldsymbol{-}$};
  \node[sum, termV] at (6.75, 5.3) (vERel) {$\boldsymbol{-}$};

  \draw[arrV] (vFnewS)-- (vFRel);
  \draw[arrV] (vMin)  .. controls(5.85, 7.5) .. (vFRel);
  \draw[arrV] (vEnewS)-- (vERel);
  \draw[arrV] (vMin)  -- (vERel);

  \node[termT] at (3, 4.8) (tF) {$\boldsymbol{t_f}$};
  \node[termT] at (5, 4.8) (tE) {$\boldsymbol{t_e}$};

  \draw[arrV] (vFRel) -- (tF);
  \draw[arrA] (aMin)  -- (tF);
  \draw[arrV] (vERel) -- (tE);
  \draw[arrA] (aMin)  -- (tE);

  \node at (-1.7, 4.5) {\small{\textbf{Iteration 3}}};
  \draw[dashed] (-2.6, 4.3) -- (8, 4.3);

  
  \node[sum, termT] at (4, 3.8) (tDif) {$\boldsymbol{-}$};
  
  \draw[arrT] (tF) -- (tDif);
  \draw[arrT] (tE) -- (tDif);

  \node[termS] at (7.5, 3.3) (sE) {$\boldsymbol{d_e}$};
  \node[termS] at (1.5, 3.3) (sF1) {$\boldsymbol{d_{f}}$};
  \node[termS] at (6, 3.3) (sF2) {$\boldsymbol{d_{fe}}$};

  \draw[arrA] (aMin)  -- (sF1);
  \draw[arrV] (vFnewS)-- (sF1);
  \draw[arrT] (tF)    -- (sF1);
  \draw[arrA] (aMin)  -- (sE);
  \draw[arrV] (vEnewS)-- (sE);
  \draw[arrT] (tE)    -- (sE);
  \draw[arrT] (tDif)  -- (sF2);
  \draw[arrV] (vMin)  -- (sF2);

  \node at (-1.7, 3) {\small{\textbf{Iteration 4}}};
  \draw[dashed] (-2.6, 2.8) -- (8, 2.8);


  \node[sum, termS] at (4, 2.3) (sF) {$\boldsymbol{+}$};
  \draw[arrS] (sF1) -- (sF);
  \draw[arrS] (sF2) -- (sF);

  \node at (-1.7, 2) {\small{\textbf{Iteration 5}}};
  \draw[dashed] (-2.6, 1.8) -- (8, 1.8);

  \node[sum, termS] at (4, 1.3) (s) {$\boldsymbol{-}$};
  \draw[arrS] (sF) -- (s);
  \draw[arrS] (sE) -- (s);

  \node at (-1.7, 1) {\small{\textbf{Iteration 6}}};
  \draw[dashed] (-2.6, 0.8) -- (8, 0.8);

  \node[sum, termS] at (0, 0.3) (sB) {$\boldsymbol{+}$};
  \draw[arrS] (s)    -- (sB);
  \draw[arrS] (dSum) -- (sB);

  \node at (-1.7, 0.3) {\small{\textbf{Iteration 7}}};
\end{tikzpicture}
	\caption{A derivation of the \enquote{can-accelerate} predicate from \autoref{eq:handcraftedPoly}.}
	\label{fig:handcraftedPredicate}
\end{figure}

In contrast to the grammar approach from~\cite{akmese}, every predicate we generate can be used directly as a splitting predicate in a decision tree -- we do not have any non-terminals we need to replace later. For example, we could try to use $d_{one} \le c$ for some constant $c\in \mathbb{R}$ in our decision tree, where we would replace the constants $a_{min}, a_{max}$ and $t_1$ in $d_{one}$ with their respective numerical values, leaving us with a function of the state variables $v_f$ and $v_e$. Or instead, we could use $d_{one}$ in the next substitution to create more sophisticated expressions.

\subsubsection{Handcrafted Predicate Derivation} \label{sec:handcraftedPredicateIntroduction}
We have seen a technique of generating compounded predicates, but how far away is the handcrafted predicate we want to synthesize? For that, we illustrate how to derive the handcrafted predicate from \autoref{eq:handcraftedPoly} using our approach.
For details, see \arxivref{Appendix~\ref{app:handcraftedPredicateIntroduction}}{\cite[Appendix B.2]{techreport}}.
In \autoref{fig:handcraftedPredicate} we show the resulting sequence of combinations that is necessary to arrive at the handcrafted predicate.

\subsubsection{Performance}
Unfortunately, the proposed approach is infeasible in practice. From 8 base identities and 9 starting values from $C \cup S$ (we leave out $a_{neutral}$), after one iteration, we already have 3,604 predicates. After the second iteration, we estimate the number of predicates to be in the realms of $10^{18}$.

An important contributor to the growth is Step 3. Without Step 3, we generate 66 predicates in the first and 10,568 in the second iteration. Unfortunately, \autoref{fig:handcraftedPredicate} shows that these sums and differences are crucial throughout all iterations of the algorithm.


The difficulties are further detailed on in \arxivref{Appendix~\ref{app:identified_problems}}{\cite[Appendix B.3]{techreport}} and lead us to  approach the problem from a different perspective in the next section.


\section{Predicates From Controller Data}\label{chapter:pred_controller_data}
\begin{figure}[ht]
  \centering
  \scalebox{0.8}{\input{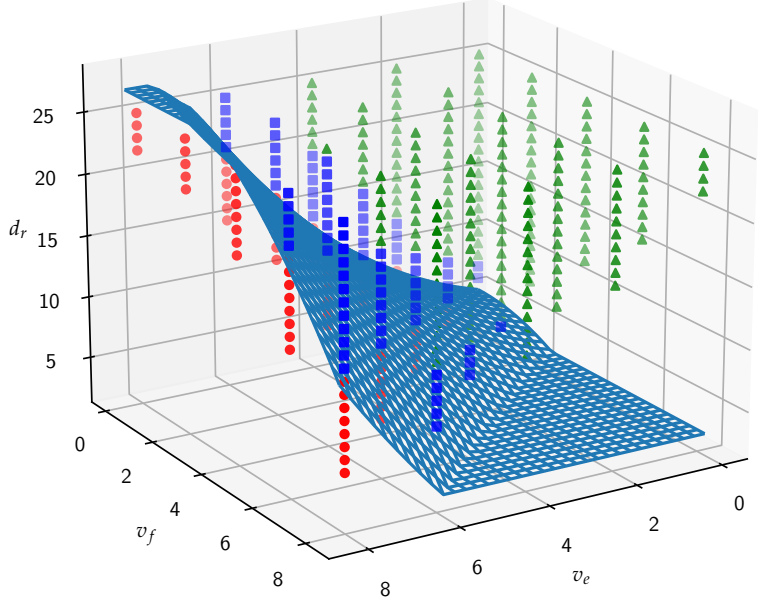}}
  \caption{Visualization of a handcrafted predicate perfectly separating the data.}
  \label{fig:plot_3d}
\end{figure}

Instead of using domain knowledge we now precisely analyze the controller data. In \autoref{fig:plot_3d}, we have visualized a part of the controller data from the \cruise\ model together with a handcrafted splitting predicate. The coordinate axes describe our three state variables $v_e, v_f, d_r$ and the color of the data points describes which actions are allowed. We see that the handcrafted strategy perfectly separates the red and blue labels. 

In this section, we use SVM to perform this task.
(For completeness, \arxivref{Appendix~\ref{sec:problem_with_curve_fitting}}{\cite[Appendix C.1]{techreport}} explores why the available curve fitting used in~\cite{weinhuber,dtcontrol2} functionality is not sufficient for our goal.) 
We apply the standard SVM learning algorithms.
These either yield linear classifiers or can use kernel methods where the resulting classifiers in principle can use, for instance, arbitrary polynomials, see \arxivref{Appendix~\ref{app:svm}}{\cite[Appendix C.2]{techreport}}.
To keep the complexity of the polynomials under control, we can restrict the degree of the polynomials.
For our running example, we know that the kinematics equations are of order two, so we can restrict to quadratic polynomials. Concretely, we change from the linear three-dimensional space $(v_e, v_f, d_r)^T$ to the quadratic space with the following 9 dimensions
\begin{align} \label{eq:highDimSpace}
	(v_e,\, v_f,\, d_r,\, v_e v_f,\, v_e d_r,\, v_f d_r,\, v_e^2,\, v_f^2,\, d_r^2)^T
\end{align}
The moderate increase in dimensions is clearly outweighed by the much better performance of the linear SVM algorithms we can now use.
Note that in this case, we used domain knowledge to estimate that no polynomials of degree more than two are necessary. Still, this requires way less manual effort than designing predicates by hand. 
Moreover, such domain knowledge is not required in general. When evaluating how the approach generalizes in Section~\ref{sec:benchmarks}, we obtain good results by always use quadratic polynomials, independent of the case study.

The polynomial predicates we receive for the \cruise\ example consist of up to 25 terms (see \arxivref{Appendix \ref{sec:prePrettifyEq}}{\cite[Appendix C.3]{techreport}} for an example). 
To make the decision trees even smaller and more explainable, we introduce four additional techniques besides the standard SVM usage. 
First, we simplify the individual predicates by removing \emph{unimportant terms} in \autoref{sec:featureImportance} and \emph{rounding the coefficients} to nice numbers in \autoref{sec:roundingCoefficients}. Then, we optimize which predicates are \emph{selected} when building the decision tree by proposing a new impurity measure in \autoref{sec:minLabelEntropy} and changing the predicates' \emph{priorities} in \autoref{sec:predPrio}.

\subsection{Feature Importance} \label{sec:featureImportance}
As we have discussed in \autoref{chapter:motivating_example}, the state of the \cruise\ model is defined by $v_e, v_f,$ and $d_r$. However, the model checker \t{UPPAAL Stratego} also exposes four additional state variables. These comprise the current acceleration values $a_e, a_f$ that do not impact the acceleration the cars choose in the next time step and the variables $f_{choose}$ and $e_{choose}$ that are an artifact from the internal model and have constant values for all relevant states.

\begin{figure}[ht]
  \centering
  \subfloat[][]{ \label{fig:feature_importance_example}
   \begin{tikzpicture}
  \begin{axis}[
    height=4.5cm,
    ticks=none,
    axis lines=center,
    xmin=0, xmax=5,
    ymin=0, ymax=5,
    xlabel=$x_1$,
    ylabel=$x_2$,
	  scatter/classes={
      a={mark=square*,blue},
      b={mark=*,red},
      c={mark=triangle*, orange}
    }
  ]
    \addplot[
      scatter,
      only marks,
      scatter src=explicit symbolic
    ]
    table[meta=label]
    {
      x   y   label
      1   1   a
      1   2   a
      1   3   a
      1   5   a
      2   1   b
      2   4   b
      3   2   c
      3   3   c
      4   1   a
      4   3   a
    };
  \end{axis}
\end{tikzpicture}
  }
  \subfloat[][]{ \label{fig:feature_importance_problem}
    \begin{tikzpicture}
  \begin{axis}[
    height=4.5cm,
    ticks=none,
    axis lines=center,
    xmin=0, xmax=5,
    ymin=0, ymax=5,
    xlabel=$x_1$,
    ylabel=$x_2$,
	  scatter/classes={
      a={mark=square*,blue},
      b={mark=*,red}
    }
  ]
    \addplot[
      scatter,
      only marks,
      scatter src=explicit symbolic
    ]
    table[meta=label]
    {
      x   y   label
      1   4   a
      1   3   a
      2   1   b
      3   2   b
      4   4   a
    };
  \end{axis}
\end{tikzpicture}
  }
  \caption{Examples with redundant features. In (a) $x_2$ is not needed. In (b) $x_1$ and $x_2$ individually look redundant, but only one may be removed. Also, removing $x_1$ is preferred over removing $x_2$}
  \label{fig:feature_importance}
\end{figure}
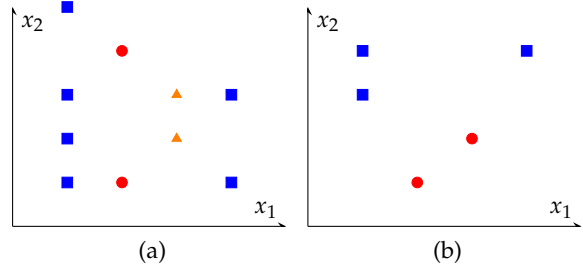

To recognize such unimportant variables, we introduce a basic version of a feature importance measure. Consider the two-dimensional dataset shown in \autoref{fig:feature_importance_example} with features $x_1$ and $x_2$. To classify a data point, feature $x_2$ is not needed. We verify this, by removing feature $x_2$ and grouping the data points with the same $x_1$ value. We can now measure how many \enquote{collisions} occur. If zero collisions occur, the feature is not needed. Otherwise, we can give a rough estimate of the importance of that feature by calculating the ratio of data points where a collision happened.

Note that for a dataset like \autoref{fig:feature_importance_problem}, this approach would judge both features as irrelevant. Individually seen that is correct but we can only remove one of them without causing collisions. This is why we calculate the feature importance incrementally. When we find an irrelevant feature, we remove it directly before calculating the importance of the next feature. As a result, the outcome may depend on the order of features we choose. For example, in \autoref{fig:feature_importance_problem}, removing $x_1$ would result in a linearly separable dataset while removing $x_2$ would not. In general, there might even be a case where we can either remove a single feature $x_i$ or all three features $x_{i+1}, x_{i+2}, x_{i+3}$. However, we did not observe any behavior like this so far, so we leave this issue for future work.

\subsection{Rounding Coefficients} \label{sec:roundingCoefficients}
With the feature importance, we remove variables that are clearly useless and reduce the number of terms in the \cruise\ predicates from 25 to 9. Still, we generate predicates that contain unnecessary terms. For example, we know from our handcrafted predicate (see \autoref{eq:handcraftedPoly}) that we do not need a $d_r^2$ term for the \cruise\ predicates. But, in the predicate we generate (see \arxivref{Appendix \ref{sec:preRounding}}{\cite[Appendix C.4]{techreport}}), the respective coefficient has a small positive value. To understand why that is the case, recall that the only objective the SVM has is to maximize the margin between the data points. For that, a small coefficient for $d_r^2$ seems to be beneficial. If we loosen the maximum margin objective, we can generate a predicate with equivalent accuracy but a simpler algebraic expression. Again, as we are not interested in the classifier's ability to generalize -- as long as the accuracy for our controller data stays the same -- we do not care about how large the margins are.\\
So, to prettify our predicate, we proceed in three steps:
\begin{enumerate}
  \item Setting coefficients to zero.
  \item Scaling the entire predicate.
  \item Rounding coefficient to integers or \enquote{nice} numbers.
\end{enumerate}

\subsubsection{Rounding to Zero}
If we can set a coefficient to zero, the predicate becomes significantly shorter and easier to understand. So this is our primary goal. A natural approach is to try setting a coefficient with a small absolute value to zero and checking if the classification for all samples stays the same.
While this suffices for some coefficients, sometimes we need to change the remaining ones to counterbalance the change. So, what we do instead is to remove the feature temporarily and try re-training the SVM. If successful, we permanently remove the feature for this split and try the next feature. Similar to the feature importance approach (\autoref{sec:featureImportance}), the result may again depend on the order of coefficients we try to remove. Here, we use the heuristic of trying to remove the coefficient with the smallest absolute value first.

Compared to the feature importance approach, three key differences make this approach more powerful:
\begin{itemize}
  \item We only consider the subset of the entire dataset available in the current subtree.
  \item We only focus on separating one specific label (we only have the two labels $+1$ and $-1$).
  \item We directly consider the features in the higher-dimensional space such as~$d_r^2$.
\end{itemize}

\subsubsection{Scaling the Predicate}
An additional step to improve readability is to scale the generated predicate. In principle, a predicate $\alpha: 0.5x + 0.1y \le 0.3$ is equivalent to a scaled predicate $10\alpha: 5x + y \le 3$ but the second one might be easier to read. The SVM uses an internal scaling constraint 
but for us, this is not relevant. We can again lift this constraint and scale all coefficients as well as the intercept value $b$ arbitrarily. One could think of various heuristic of how to scale the predicate. We decided to use a simple one: we search for the coefficient with the value closest to 1 and scale the predicate so that it becomes exactly $1$. This way, we have at least one term with a simple coefficient.

\subsubsection{General Rounding}
As the last step, we generalize the \enquote{rounding to zero} approach and use it on the coefficient we could not set to zero. This way, instead of having a predicate like $8.165839d_r^2 - 2.935846v_r \le 0$ we can use a nicer looking one like $8d_r^2 - 3v_r \le 0$. For that, we try the approach from above with increasing relative precision. For example, for the coefficient of $d_r^2$, we first try the value $10$, then $8$, then $8.2$, and so on, until we find a value that does not change the classification for any sample. Note that we do not re-train the SVM in this step but simply change the coefficient and check if the classification stays the same.

\paragraph{}
With these techniques, we can finally generate pretty predicates. For example, one the predicate we find for the \t{cruise\_250} dataset exactly corresponds to the handcrafted polynomial from \autoref{eq:handcraftedPoly} after substituting all constants. The only difference is the constant offset.
\begin{align}
  -0.25v_e^2 + 0.25v_f^2 -5v_e+3v_f + d_r +  19.5 \le 0
\end{align}
These rounding procedures raise a number of technical numerical issues, but we describe ways to mitigate these in \arxivref{Appendix~\ref{app:num}}{\cite[Appendix C.5]{techreport}}.

\subsection{Min-label Entropy} \label{sec:minLabelEntropy}
\begin{figure}[ht]
  \centering
\includegraphics{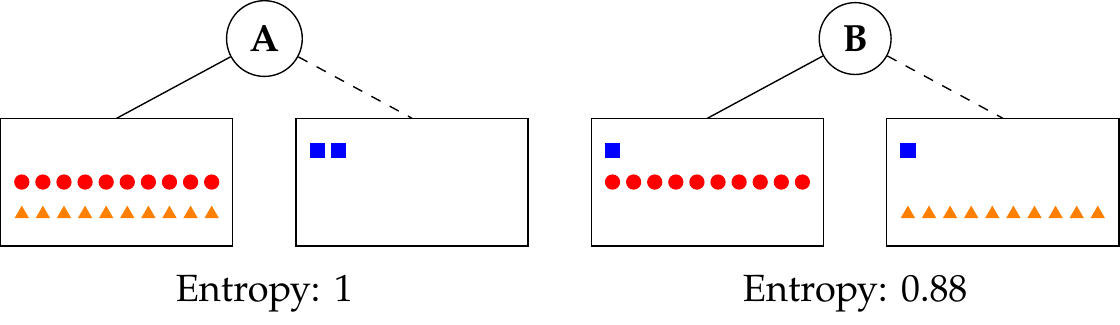}
  \caption{Two different splits with their respective entropy values. While split B has a better entropy value and is preferred in machine learing, we want to use split A first when building a perfect classifier.}
  \label{fig:entropy_split}
\end{figure}

Now that we have pretty predicates, we shift our focus to the decision tree construction for the next two sections. For the \cruise\ dataset, we can now construct a decision tree with 37 nodes, only 10\% of the size when using linear predicates. Moreover, we have seen that the approach generates the exact predicates we derived by hand in \arxivref{Appendix~\ref{app:hand}}{\cite[Appendix A.3]{techreport}}. Still, the decision tree is not as compact as the 11 node tree from \cite{akmese} as we do not directly use those predicates. To understand why this is the case, we have a look at \autoref{fig:entropy_split}. We see that split A perfectly separates the blue label from the rest, while B separates the red and orange labels but distributes the blue one among both children. Considering only a single split, we would prefer split B because the dataset is nicely separated except for the small number of blue samples. The entropy impurity measure comes to the same conclusion and assigns split B a better entropy score.

However, when building a perfect classifier for representing the most-permissive controller, we have a different perspective than in machine learning. At some point, we need to separate the blue labels from the rest. If we do not separate them now and select split B, we have to add additional splits on \i{both sides} of the split B. If we rather start with split A, we can select split B as the next split in the left child and thus receive a smaller decision tree. 


This effect is especially prevalent if the number of samples per label differs significantly. In the \cruise\ example, we observe exactly that: the label \enquote{all actions are allowed} has 20 times more data points than any of the other labels. Hence, we introduce a new impurity measure that we call \i{min-label entropy}:

\begin{definition}
  For a dataset $X = X_l \uplus X_r$ with the label set $B$, let $n(X, y)$ describe the number of data points in $X$ with label $y \in B$. 
  For a predicate $\alpha$ that splits the dataset into $X_l$ and $X_r$, we define the \i{min-label entropy} $H^*$ as 
  \begin{align}
    K(p)           &\mathrel{:=} -p \log_2(p) \nonumber\\
    H^*(X, y)      &\mathrel{:=} K\left(\frac{n(X, y)}{|X|}\right) \nonumber\\ 
    H^*(\alpha ,X) &\mathrel{:=} \min_{y \in B} \left\{ \frac{|X_l|}{|X|} H^*(X_l, y) + \frac{|X_r|}{|X|} H^*(X_r, y)\right\}
  \end{align}
\end{definition}
Intuitively, the \i{min-label entropy} measure estimates for every label $y$, how difficult it will be to to separate the label $y$ in both partions after this split. Then it returns the value of the best label. The strategy we want to provoke with this impurity measure is to first fully separate one label and then continue with the next one. Specifically, if we can completely separate one label like in the example in \autoref{fig:entropy_split}, the impurity for this split is 0 and we definitely select such a spilt.

\subsection{Predicate Priority} \label{sec:predPrio}
With the min-label entropy, we reduce the decision tree size of the \cruise\ example to 25. As a last optimization heuristic, we also adjust the priorities of the predicates. When deciding between an axis-aligned and a polynomial predicate that both have similar impurity values, we want to choose the axis-aligned one as it is considerably simpler to understand. For that reason, \dtControl\ has implemented a priority function for predicate generators. For example, when we give the polynomial predicates the priority $0.5$ and the axis-aligned ones the priority $1$, we only choose a polynomial predicate if it is at least twice as good in terms of the impurity measure. In fact, we want to choose an even lower value as a priority for another reason. In the \cruise\ example, we know that we can find a polynomial that distinguishes cases where we can accelerate from those where we cannot. In our handpicked strategy, we did however not consider the edge cases when we are already driving at minimal or maximal velocity. If we do not exclude those, the data is not perfectly separable, meaning we will find a polynomial split that almost classifies everything correctly, but misses a few data points. While this is not a huge problem, it turns out that it is more effective to first exclude the edge cases with axis-aligned predicates and then perfectly split the data with a complex predicate later. We can achieve it with a low priority value $\le 0.2$ for the polynomial splits in combination with our min-label entropy. This way, we will only choose the complicated splits if they are at least 5 times better. Note that the impurity is $0$ if we can perfectly separate one label, so in this case, we are infinitely better than any non-perfect solution.


\section{Experimental Evaluation}\label{chapter:evaluation}
In this section, we will evaluate our approach experimentally.
As we did not succeed in creating truly explainable decision trees for the cruise example using the method proposed in \autoref{chapter:pred_domain_knowledge}, we focus on the method of \autoref{chapter:pred_controller_data}. 
For completeness, in \arxivref{Appendix \ref{app:exp-domain}}{\cite[Appendix D.1]{techreport}} we provide the evaluation of the approach described in \autoref{chapter:pred_domain_knowledge} .
Thus we evaluate how well generating quadratic polynomials with SVMs performs in practice. While developing the various techniques and heuristics, we mainly focused on the \cruise\ dataset. In this section, we first analyze the results for this dataset but then investigate how well the approach generalizes to other case studies. For that, we compare our results to the existing approaches and to the minimum decision tree achievable in theory. Afterwards, we look at our new impurity measure and its performance independently and finally comment on explainability.

\paragraph{Artifacts}
All resources such as generated domain knowledge predicates, model files, and synthesized controllers used in this paper are available to download at \cite{thesisFiles}. The repository also contains scripts to reproduce the benchmark tables presented in this paper.

\subsection{Cruise Control}
\begin{figure}[t]
  \centering
  \subfloat[][\t{cruise\_250}]{
    \resizebox{!}{9.5cm}{\includegraphics{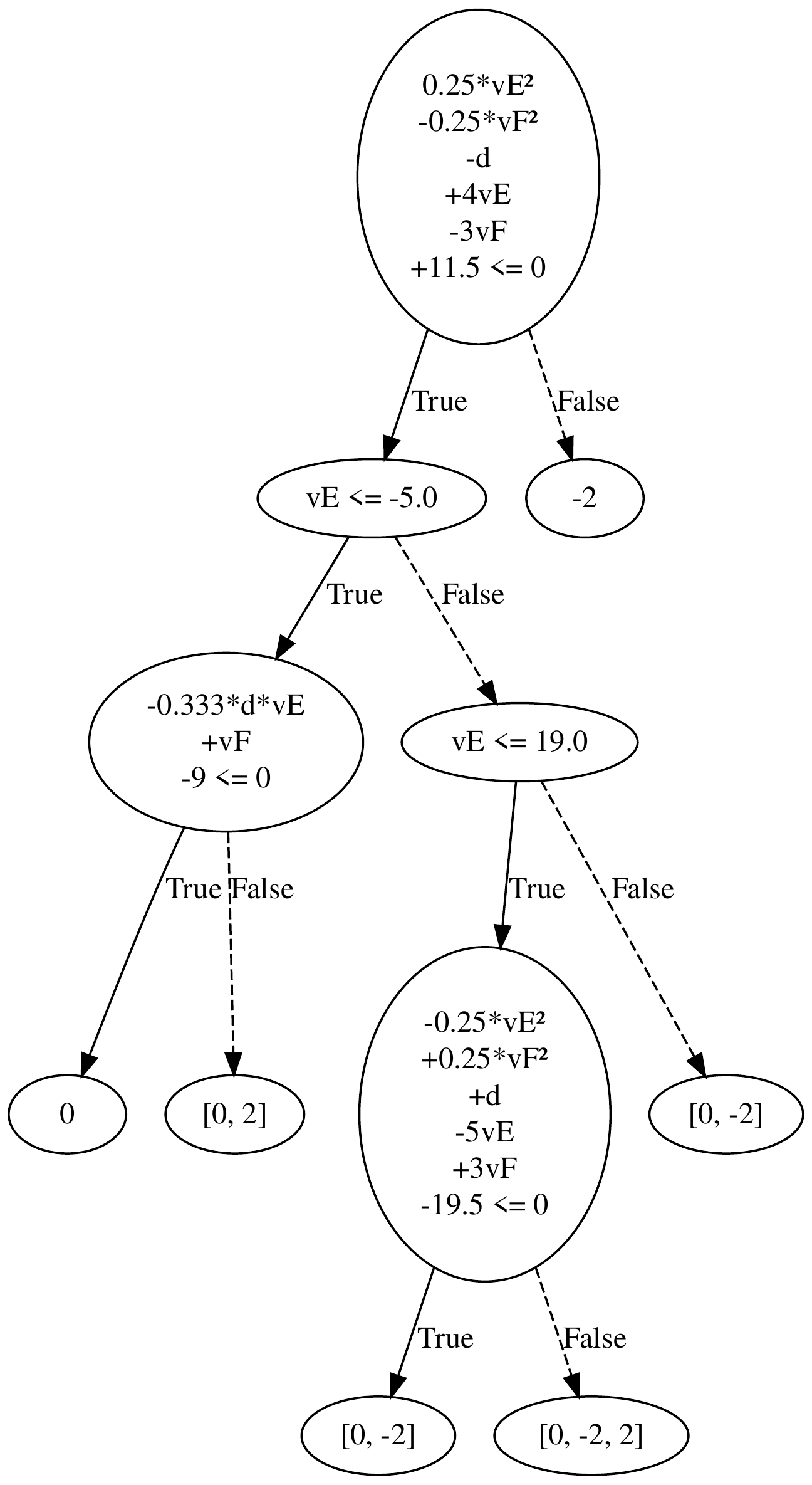}}
    \label{fig:dt_cruise_250}
  }
  \subfloat[][\t{cruise\_300}]{
    \resizebox{!}{9.5cm}{\includegraphics{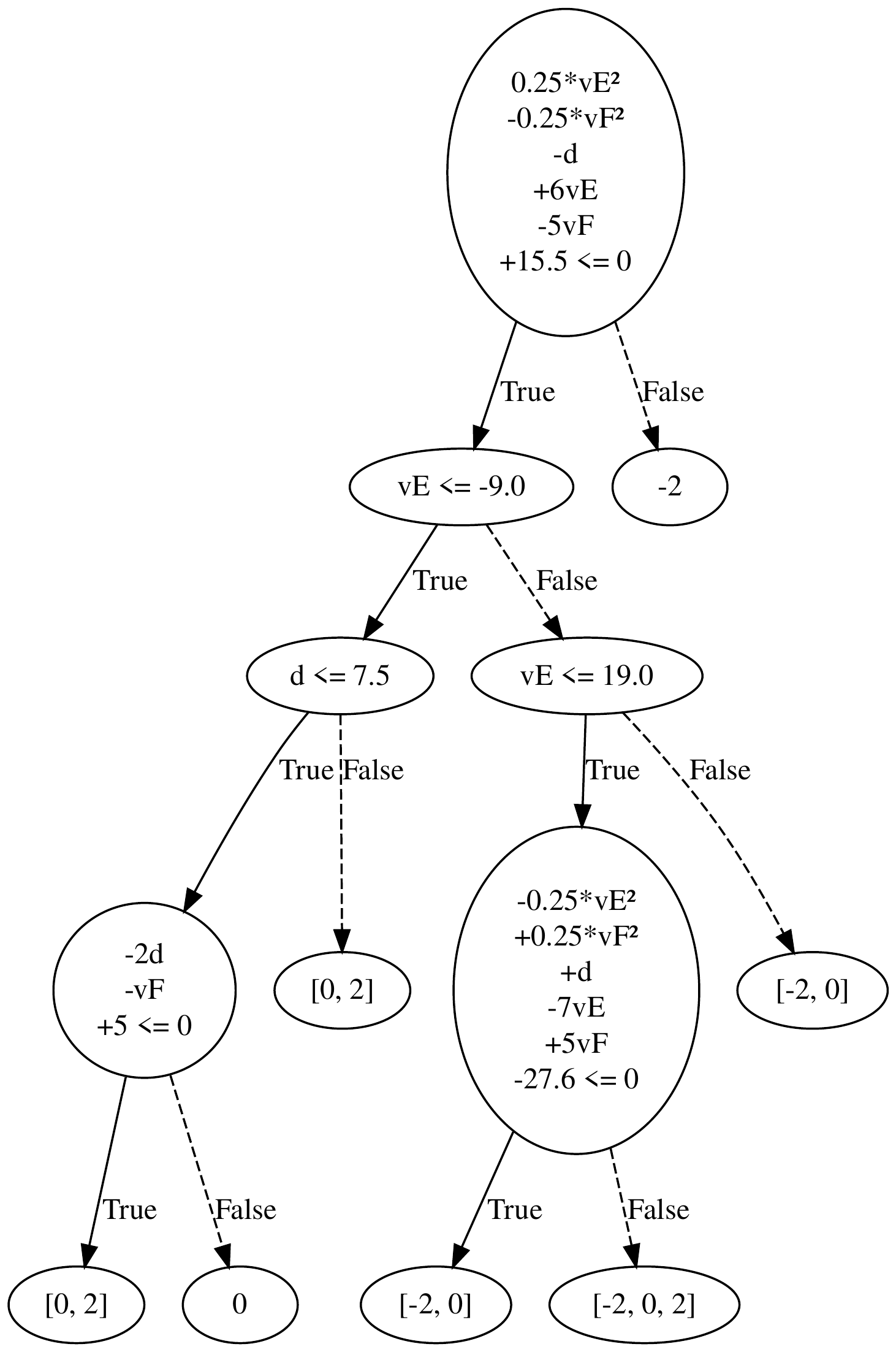}}
    \label{fig:dt_cruise_300}
  }
  \caption{The decision trees for the \cruise\ example generated by our data-driven approach.}
  \label{fig:dt_cruise}
\end{figure}

Using all the strategies discussed in \autoref{chapter:pred_controller_data} we achieve great results for the \cruise\ model. For the \t{cruise\_250} dataset, we find a succinct decision tree with only 11 nodes (see \autoref{fig:dt_cruise_250}). This is exactly the number of nodes \cite{akmese} found with the handcrafted strategy. In fact, we precisely found the handcrafted \enquote{must break} and \enquote{can accelerate} predicate from the handcrafted strategy, with only a small difference in the constant offset.

For the slightly larger \t{cruise\_300} dataset, we generate a very similar but slightly larger decision tree with 13 nodes (\autoref{fig:dt_cruise_300}). The quadratic predicates change in line with the change of the constant $v_{min}$ (see \arxivref{\autoref{tab:cruise_parameters} in the Appendix~\ref{sec:cruiseParameters}}{see Table 3 in \cite[Appendix A.2]{techreport}}) and one complex splitting predicate is exchanged for two simpler predicates.

\medskip
In both cases, the generated decision trees are almost 80 times smaller than the ones we receive with axis-aligned predicates and 30 times smaller than the ones with linear predicates.

\subsection{Generalizing to Other Benchmarks} \label{sec:benchmarks}

\begin{table}
	\centering
	\small
	\caption{The number of nodes of the generated decision trees using axis-aligned splits, linear splits, and the proposed quadratic polynomial splits with priority 0.1 and 1.0.
		Each row displays the result using the entropy impurity measure at the top and using min-label entropy at the bottom. TO means time out after 3 hours.
		As a comparison, we show the number of states of the underlying controller and the theoretical minimum size a decision tree needs to have (see \arxivref{Appendix~\ref{app:min}}{\cite[Appendix D.2]{techreport}}). The full table is in \arxivref{Appendix \ref{chp:all_results}}{\cite[Appendix D.3]{techreport}}.
	}
	\begin{tabular}{@{}lrr rrrr@{}} \toprule
  & \multicolumn{2}{c}{Comparision}
& \multicolumn{2}{c}{Previous} & \multicolumn{2}{c}{Quadratic} \\
\cmidrule(r){2-3} \cmidrule(r){4-5} \cmidrule(r){6-7}
Case Study & States & MinSize
& Ax.Al. & Linear & Poly & PolyPrio1\\ \midrule
\multirow{2}{*}[0pt]{cartpole \cite{Jagtap2020}}
& \multirow{2}{*}[0pt]{271} & \multirow{2}{*}[0pt]{\b{169}}
  & 253    & 183   & 243  & 189 \\
&&& 263    & 187  & \b{169}  & \b{169} \\\addlinespace[0.4em]

\multirow{2}{*}[0pt]{10rooms \cite{quest}}
& \multirow{2}{*}[0pt]{26,244} & \multirow{2}{*}[0pt]{\b{49}}
& 17,297 & 147       & 61   & 61 \\
&&& 17,297 & 107    & \b{49}   & \b{49} \\\addlinespace[0.4em]

\multirow{2}{*}[0pt]{helicopter \cite{Jagtap2020}}
& \multirow{2}{*}[0pt]{280,539} & \multirow{2}{*}[0pt]{475}
& 6,339  & \b{3,769} & 5,035& 3,787 \\
&&& 9,649 & 4,637& TO  & TO \\\addlinespace[0.4em]

\multirow{2}{*}[0pt]{cruise\_250 \cite{cruise}} 
& \multirow{2}{*}[0pt]{320,523} & \multirow{2}{*}[0pt]{9}
& 869     & 369   & 353  & 37 \\
&&& 1,067  & 363   & \b{11}   & 25 \\\addlinespace[0.4em]


\multirow{2}{*}[0pt]{dcdc \cite{scots}}
& \multirow{2}{*}[0pt]{593,089} & \multirow{2}{*}[0pt]{5}
& 271    & 139  & \b{129}  & 199 \\
&&& 265     & 173   & 147  & 273 \\\addlinespace[0.4em]

\multirow{2}{*}[0pt]{truck\_trailer \cite{pfaces}} 
& \multirow{2}{*}[0pt]{1,386,211} & \multirow{2}{*}[0pt]{1,839}
& \b{338,283}& TO   & TO  & TO \\
&&& 366,411 & TO   & TO  & TO \\

\multirow{2}{*}[0pt]{aircraft \cite{Rungger2015}}
& \multirow{2}{*}[0pt]{2,135,056} & \multirow{2}{*}[0pt]{31}
& 915,877& 916,685 & 725,011& \b{602,335} \\
&&&1,015,903&1,013,949& 688,577& 630,631 \\


\midrule

\multirow{2}{*}[0pt]{pacman.5} & \multirow{2}{*}[0pt]{232}& \multirow{2}{*}[0pt]{\b{37}}
&      53&      49&      47&      \b{37}\\
&&&      81&      59&      \b{37}&      \b{37}\\\addlinespace[0.4em]
\multirow{2}{*}[0pt]{philosophers-mdp.3} & \multirow{2}{*}[0pt]{344}& \multirow{2}{*}[0pt]{59}
&     391&     333&     315&     251\\
&&&     403&     367&     251&     \b{223}\\\addlinespace[0.4em]
\multirow{2}{*}[0pt]{ij.10} & \multirow{2}{*}[0pt]{1,013}& \multirow{2}{*}[0pt]{19}
&   1,291&     753&     897&     209\\
&&&   1,405&     735&     \b{141}&     177\\\addlinespace[0.4em]
\multirow{2}{*}[0pt]{elevators.a-11-9} & \multirow{2}{*}[0pt]{14,742}& \multirow{2}{*}[0pt]{129}
&  16,341&   9,865& 9,779 & 2,859\\
&&&  17,809&   9,955& 2,023 & \b{1,919} \\\addlinespace[0.4em]
\multirow{2}{*}[0pt]{exploding-blocksworld.5} & \multirow{2}{*}[0pt]{76,741}& \multirow{2}{*}[0pt]{149}
&  16,913&   2,687&   4,511&     \b{829}\\
&&&  20,273&   2,845&TO&TO\\\addlinespace[0.4em]
\multirow{2}{*}[0pt]{wlan\_dl.0.80.deadline} & \multirow{2}{*}[0pt]{189,641}& \multirow{2}{*}[0pt]{175}
&   3,369&     701&     693&     667\\
&&&   3,675&   2,841& \b{523} &TO\\\addlinespace[0.4em]
\multirow{2}{*}[0pt]{pnueli-zuck.5} & \multirow{2}{*}[0pt]{303,427}& \multirow{2}{*}[0pt]{173}
& 171,371& 156,165& 114,979&  \b{83,219}\\
&&& 263,955& 221,645&  95,879& 83,951\\

\bottomrule
\end{tabular}
	\label{tab:results_condensed}
\end{table}

To see how our approach and the individual heuristics generalize, we evaluate them on the case studies of cyber-physical systems from \cite{dtControl} as well as on the case studies from the quantitative verification benchmark set \cite{Hartmanns2019} that were used in \cite{dtcontrol2}. We avoid using any determinization heuristics so that we generate the most-permissive controllers. We ran all experiments on a server with the operating system Ubuntu 20.04, a 2.2GHz Intel(R) Xeon(R) CPU E5-2630 v4 and 250 GB RAM. \autoref{tab:results_condensed} contains a selection of the results, with the case studies of cyber-physical systems at the top and quantitative verification at the bottom. In every row, we compare the number of nodes in the generated decision tree for
\begin{itemize}
  \item the axis-aligned splitting strategy (Ax.Al.),
  \item the smallest decision tree we could generate with axis-aligned and linear predicates\footnote{We calculate this as the minimum over the three splitting strategies logistic regression, linear support vector machines, and the OC1 heuristic.} (Linear),
  \item axis-aligned predicates and the quadratic polynomials generated by support vector machines with a priority value of 0.1 (Poly),
  \item and with the default priority value of 1.0 (PolyPrio1).
\end{itemize}

In every cell, the top number describes the result using the entropy impurity measure and the bottom number refers to the result using min-label entropy. TO indicates that we were not able to generate a decision tree within three hours.
As a comparison, we list the number of states of the controller as well as the theoretical minimum size of the decision tree (see \arxivref{Appendix~\ref{app:min}}{\cite[Appendix D.2]{techreport}} for a description of how to calculate this).

A complete table with all 28 case studies, a comparison with BDDs, and results for different linear strategies can be found in \arxivref{Appendix \ref{chp:all_results}}{\cite[Appendix D.3]{techreport}}.

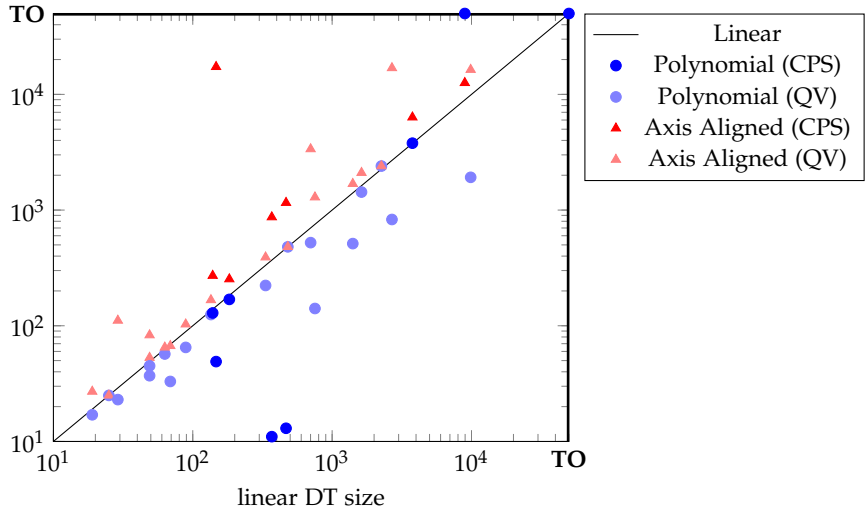
\begin{figure}[t]
	\centering
	\begin{tikzpicture}
  \begin{axis}[
    ymin=10,
    xmin=10,
    ymax=50000,
    xmax=50000,
    xmode=log,
    ymode=log,
    xlabel={linear DT size},
    extra y ticks={50000},
    extra y tick labels={\textbf{TO}},
    extra x ticks={50000},
    extra x tick labels={\textbf{TO}},
    legend pos=outer north east,
	  scatter/classes={
      A={mark=*,blue},
      a={mark=*,blue!50},
      C={mark=triangle*,red},
      c={mark=triangle*,red!50}
    },
  ]
    \legend{Linear, Polynomial (CPS), Polynomial (QV), Axis Aligned (CPS), Axis Aligned (QV)}
    \addplot [
      domain=10:50000,
      color=black,
    ] {x};
    \addplot[
      scatter,
      only marks,
      scatter src=explicit symbolic
    ]
    table[meta=label]
    {
      x        y label
      19      17 a
      19      27 c
      49      37 a
      49      53 c
     481     481 a
     481     481 c
     333     223 a
     333     391 c
      25      25 a
      25      25 c
      29      23 a
      29     111 c
     753     141 a
     753    1291 c
      49      45 a
      49      83 c
    1407     513 a
    1407    1687 c
    2257    2401 a
    2257    2401 c
      69      33 a
      69      67 c
      63      57 a
      63      65 c
      89      65 a
      89     103 c
     135     125 a
     135     167 c
    9865    1919 a
    9865   16341 c
    2687     829 a
    2687   16913 c
    1625    1431 a
    1625    2101 c
     701     523 a
     701    3369 c
     183     169 A
     183     253 C
     147      49 A
     147   17297 C
    3769    3787 A
    3769    6339 C
     369      11 A
     369     869 C
     467      13 A
     467    1157 C
     139     129 A
     139     271 C
   50000   50000 A
    8953   50000 A
    8953   12573 C
    };
   
    \draw[ultra thick] (0, 50000) -- (50000, 50000);
    \draw[ultra thick] (50000, 0) -- (50000, 50000);
  \end{axis}
\end{tikzpicture}
	\caption{Performance comparison of different predicate types. Based on the decision tree size using linear predicates, we compare how many nodes the decision trees with axis-aligned splits and quadratic polynomials have. Every sample corresponds to a case study of cyber-physical systems (CPS) or originates from the quantitative verification benchmark set (QV).}
	\label{fig:performance_scatter}
\end{figure}

\paragraph{Scatter Plot}
To complement the table, \autoref{fig:performance_scatter} visualizes the results in a logarithmic scatter plot. As a reference, we take the smallest tree we could generate with linear predicates and the entropy impurity. Then we compare it to the size of the tree with axis-aligned predicates and our quadratic polynomials. For example, the two blue points near the location $(370, 10)$ are the two cruise datasets. The $x$-coordinate is the size of the tree with linear predicates and the $y$-coordinate shows the size of the polynomial or axis-aligned results.

\paragraph{Analysis}
Our new approach gives smaller decision trees for almost all case studies, except for \t{helicopter} and \t{cdrive.10}, where the linear solution is smaller by 6\% and \t{traffic\_30m} where we run into a timeout (see the full table of results in \arxivref{Appendix \ref{chp:all_results}}{\cite[Appendix D.3]{techreport}}). In \autoref{tab:results_statistics} we show the cumulated statistics. Most notably, the number of cases where we find a tree of minimum size has increased from 2 to 10 out of 28.

\begin{table}
  \centering
  \caption{Cumulated statistics over all \textbf{28} benchmarks. We compare the best linear strategy with entropy impurity with the best of our heuristics.}
  \begin{tabular}{lrr} \toprule
             & Linear   & Quadratic \\
  \midrule
  Timeout    & 1       & 2\\
  Minimal DT &  \b{2} (7\%) & \b{10} (35\%)\\\addlinespace[0.5em]
  \multicolumn{2}{l}{DT is smaller or equal} & \b{25} (89\%) \\
  \multicolumn{2}{l}{DT has less than half the size} & \b{8} (29\%)\\
  \bottomrule
\end{tabular}
  \label{tab:results_statistics}
\end{table}

\subsection{Min-Label Entropy and Predicate Priority} \label{sec:eval_mle}
We applied two significant changes to arrive at the small decision trees in the \cruise\ example: the min-label entropy impurity measure and the modified predicate priority value. We now analyze how useful they are on their own for the other case studies.

\begin{figure}[ht]
  \centering
  \begin{tikzpicture}
  \begin{axis}[
    ymin=10,
    xmin=10,
    ymax=50000,
    xmax=50000,
    xmode=log,
    ymode=log,
    xlabel={Entropy Prio. 1 DT size},
    extra y ticks={50000},
    extra y tick labels={\textbf{TO}},
    extra x ticks={50000},
    extra x tick labels={\textbf{TO}},
    legend pos=outer north east,
	  scatter/classes={
      ME={mark=*,blue},
      me={mark=*,blue!50},
      LP={mark=triangle*,red},
      lp={mark=triangle*,red!50}
    },
  ]
    \legend{Entropy Prio. 1, MLE Prio. 1 (CPS), MLE Prio. 1 (QV), Entropy Prio. 0.1 (CPS), Entropy Prio. 0.1 (QV)}
    \addplot [
      domain=10:50000,
      color=black,
    ] {x};
    \addplot[
      scatter,
      only marks,
      scatter src=explicit symbolic
    ]
    table[meta=label]
    {
      x        y label
      17      17 me
      17      25 lp
      37      37 me
      37      47 lp
     481     481 lp
     481     481 me
     251     223 me
     251     315 lp
      25      25 lp
      25      25 me
      27      23 me
      27      69 lp
     209     177 me
     209     897 lp
      57      45 me
      57      75 lp
     891     513 me
     891    1451 lp
    2401    2401 lp
    2401   50000 me
      51      33 me
      51      57 lp
      59      57 me
      59      65 lp
      79      65 me
      79     103 lp
     141     125 me
     141     133 lp
    2859    1919 me
    2859    9779 lp
     829    4511 lp
     829   50000 me
    1431    2005 lp
    1431   50000 me
     667     693 lp
     667   50000 me 
     189    169 ME
     189    243 LP
      61     49 ME
      61     61 LP
    3787  50000 ME
    3787   5035 LP
      37     25 ME
      37    353 LP
      59     23 ME
      59    521 LP
     199    273 ME
     199    129 LP
   50000  50000 ME
   50000  50000 LP
    };
   
    \draw[ultra thick] (0, 50000) -- (50000, 50000);
    \draw[ultra thick] (50000, 0) -- (50000, 50000);
  \end{axis}
\end{tikzpicture}
  \caption{Performance comparison of the min-label entropy (MLE) and the low priority heuristic. Based on the decision tree size using quadratic polynomials as predicates with entropy and priority 1.0, we compare how the heuristics change the tree size. Every sample corresponds to a case study of cyber-physical systems (CPS) or originates from the quantitative verification benchmark set (QV).}
  \label{fig:performance_scatter_minEntropy}
\end{figure}
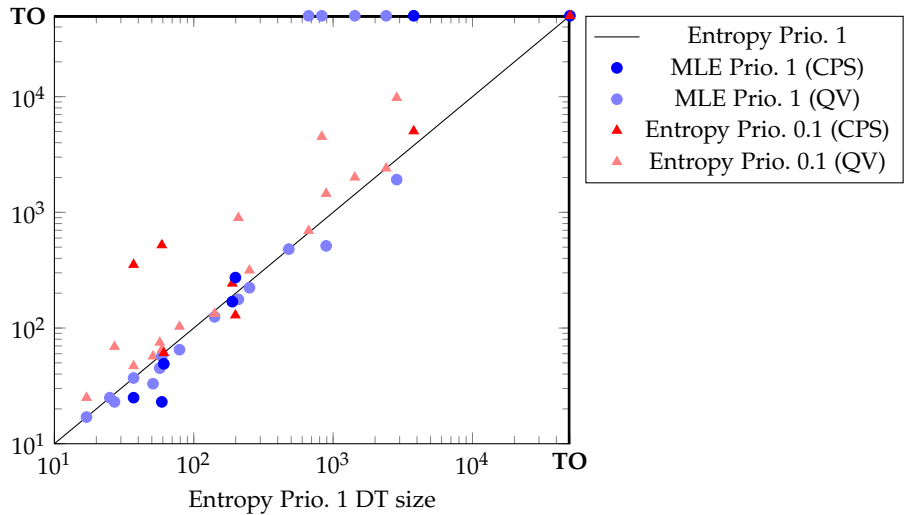

In \autoref{fig:performance_scatter_minEntropy} we again make use of a logarithmic scatter plot to visualize the data from our tables. As a baseline, we take the size of the decision tree generated with our proposed approach using the entropy impurity measure and the default priority 1.0. We compare it to the size when using the proposed min-label entropy (blue) and when using the reduced priority value 0.1 (red).

\paragraph{Min-Label Entropy} 
The min-label entropy reduces the tree size in 14 out of 17 cases (82\%) where we are not already at the minimum size and do not run into a timeout. Interestingly, this behavior is different when using the min-label entropy with axis-aligned splits or linear splits. There, the min-label entropy can only improve the result in 30 out of 106 cases (28\%).

Also, we observe 5 cases where our approach only times out when using the min-label entropy but not when using the standard entropy. A reason for this might be that the min-label entropy encourages the formation of decision trees formed like a line. For all case studies where we generate minimum-sized trees like the \t{10rooms} case study, every leaf has a unique label. With the min-label entropy impurity, every splitting predicate separates out one of those labels. So the tree looks like a line. As a consequence, the runtime for finding predicates does not decrease as fast while constructing the tree. When we construct a perfectly balanced tree, the size of the dataset left at the subtree at depth $d$ is only a small fraction ($2^{-d}$) of the original size. In the case of a line, however, the dataset size only decreases slowly.

\paragraph{Low Priority Heuristic}
While the low priority value helps in the cruise example in combination with the min-label entropy, the only other cases where this heuristic brings an improvement are the \t{dcdc} and \t{eajs.2.100.5.ExpUtil} case studies (see the full table of results in \arxivref{Appendix \ref{chp:all_results}}{\cite[Appendix D.3]{techreport}}). We conclude that our motivating idea of first separating the \enquote{outliers} and then using the more sophisticated splits later does not generalize well. Apparently, it is beneficial to just take the best available split right away in complex models.

\subsection{Explainability}
We have seen that we can significantly reduce the number of decision tree nodes with our proposed approach. But how explainable are the trees we generate?

Of course, reducing the number of decision nodes already helps create an explainable decision tree. Still, we have to consider that the complexity of the individual splitting predicate increases, thereby potentially reducing explainability. As an example, we consider the \t{10rooms} case study. Here, we find a decision tree with 49 nodes which is the minimum size for a most-permissive decision tree. 
Unfortunately, the decision tree is not particularly explainable as some predicates comprise up to 35 terms, even after trying to round coefficients to zero. The reason is the large number of 10 state variables. A quadratic polynomial with ten variables can already have 65 terms.


Regardless of the complexity of individual predicates, for some case studies, the minimum decision tree size is already too large to be easily understandable by a human. Any most-permissive decision tree for the case studies \t{helicopter} and \t{truck\_trailer} will have more than 400 and 1,800 nodes respectively. So, in these cases, we might need to investigate determinized controllers as discussed in \cite{dtControl,dtcontrol2}.

\section{Conclusion}\label{chapter:conclusion}
In this work, we have investigated two approaches to generating expressive algebraic splitting predicates for decision trees. 
We have seen that automatically generating predicates from domain knowledge is not yet feasible with the current method. Hence we proposed learning quadratic polynomials with support vector machines directly from the controller data. Additionally, we introduced a new impurity measure called \i{min-label entropy} that focuses on separating one specific label first. We integrated both ideas into the decision-tree learning algorithm and implemented it in the open-source tool \dtControl. We were able to generate significantly smaller decision trees in cases where the determinization heuristics could not be applied.
For the cruise model, we generated a tree with the same size as the one created with help of a human expert, and in 10 out of 28 case studies, we even found a decision tree of minimum size.

On the one hand, we showed that more expressive quadratic polynomials can help to generate succinct trees for permissive controllers. On the other hand, a key aspect still to be improved upon is the explainability. Of course, succinct decision trees are already easier to understand by nature, but more complex predicates again reduce explainability. Ideally, we would want to automatically generate a justification explaining the coefficients of each complex predicate. 

\bibliographystyle{alpha} 
\bibliography{bibliography}

\newcommand{\etalchar}[1]{$^{#1}$}
\begin{thebibliography}{PWH{\etalchar{+}}04}

\bibitem[ABC{\etalchar{+}}19]{Ashok2019}
Pranav Ashok, Tom{\'a}{\v{s}} Br{\'a}zdil, Krishnendu Chatterjee, Jan K{\v
  r}et{\'i}nsk{\'y}, Christoph~H. Lampert, and Viktor Toman.
\newblock Strategy representation by decision trees with linear classifiers.
\newblock In David Parker and Verena Wolf, editors, {\em Quantitative
  Evaluation of Systems}, pages 109--128, Cham, 2019. Springer International
  Publishing.

\bibitem[AJJ{\etalchar{+}}20]{dtControl}
Pranav Ashok, Mathias Jackermeier, Pushpak Jagtap, Jan K\v{r}et\'{\i}nsk\'{y},
  Maximilian Weininger, and Majid Zamani.
\newblock Dtcontrol: Decision tree learning algorithms for controller
  representation.
\newblock In {\em Proceedings of the 23rd International Conference on Hybrid
  Systems: Computation and Control}, HSCC '20, New York, NY, USA, 2020.
  Association for Computing Machinery.

\bibitem[AJK{\etalchar{+}}21]{dtcontrol2}
Pranav Ashok, Mathias Jackermeier, Jan Kret{\'{\i}}nsk{\'{y}}, Christoph
  Weinhuber, Maximilian Weininger, and Mayank Yadav.
\newblock dtcontrol 2.0: Explainable strategy representation via decision tree
  learning steered by experts.
\newblock In {\em {TACAS} {(2)}}, volume 12652 of {\em Lecture Notes in
  Computer Science}, pages 326--345. Springer, 2021.

\bibitem[AKL{\etalchar{+}}19]{sos}
Pranav Ashok, Jan Kret{\'{\i}}nsk{\'{y}}, Kim~Guldstrand Larsen, Adrien~Le
  Co{\"{e}}nt, Jakob~Haahr Taankvist, and Maximilian Weininger.
\newblock {SOS:} safe, optimal and small strategies for hybrid markov decision
  processes.
\newblock In David Parker and Verena Wolf, editors, {\em Quantitative
  Evaluation of Systems, 16th International Conference, {QEST} 2019, Glasgow,
  UK, September 10-12, 2019, Proceedings}, volume 11785 of {\em Lecture Notes
  in Computer Science}, pages 147--164. Springer, 2019.

\bibitem[Akm19]{akmese}
Safa~Mert Akmese.
\newblock Generating richer predicates for decision trees.
\newblock Bachelor's thesis, Technical University of Munich, 2019.

\bibitem[Arl94]{curvefittingBook}
S.~Arlinghaus.
\newblock {\em Practical Handbook of Curve Fitting}.
\newblock Taylor \& Francis, 1994.

\bibitem[BB98]{Bennett1998}
K.~P. {Bennett} and J.~A. {Blue}.
\newblock A support vector machine approach to decision trees.
\newblock In {\em 1998 IEEE International Joint Conference on Neural Networks
  Proceedings. IEEE World Congress on Computational Intelligence (Cat.
  No.98CH36227)}, volume~3, pages 2396--2401 vol.3, 1998.

\bibitem[BCC{\etalchar{+}}15]{counterexample}
Tom{\'{a}}s Br{\'{a}}zdil, Krishnendu Chatterjee, Martin Chmelik, Andreas
  Fellner, and Jan Kret{\'{\i}}nsk{\'{y}}.
\newblock Counterexample explanation by learning small strategies in markov
  decision processes.
\newblock In Daniel Kroening and Corina~S. Pasareanu, editors, {\em Computer
  Aided Verification - 27th International Conference, {CAV} 2015, San
  Francisco, CA, USA, July 18-24, 2015, Proceedings, Part {I}}, volume 9206 of
  {\em Lecture Notes in Computer Science}, pages 158--177. Springer, 2015.

\bibitem[BFG{\etalchar{+}}97]{ADD}
R.~Iris Bahar, Erica~A. Frohm, Charles~M. Gaona, Gary~D. Hachtel, Enrico Macii,
  Abelardo Pardo, and Fabio Somenzi.
\newblock Algebraic decision diagrams and their applications.
\newblock {\em Formal Methods Syst. Des.}, 10(2/3):171--206, 1997.

\bibitem[BFOS84]{cart}
Leo Breiman, J.~H. Friedman, R.~A. Olshen, and C.~J. Stone.
\newblock {\em Classification and Regression Trees}.
\newblock Wadsworth, 1984.

\bibitem[Bry86]{bdd}
Randal~E. Bryant.
\newblock Graph-based algorithms for boolean function manipulation.
\newblock {\em {IEEE} Trans. Computers}, 35(8):677--691, 1986.

\bibitem[BW96]{Bollig1996}
Beate Bollig and Ingo Wegener.
\newblock Improving the variable ordering of obdds is np-complete.
\newblock {\em {IEEE} Trans. Computers}, 45(9):993--1002, 1996.

\bibitem[CHC{\etalchar{+}}10]{Chang2010}
Yin{-}Wen Chang, Cho{-}Jui Hsieh, Kai{-}Wei Chang, Michael Ringgaard, and
  Chih{-}Jen Lin.
\newblock Training and testing low-degree polynomial data mappings via linear
  {SVM}.
\newblock {\em J. Mach. Learn. Res.}, 11:1471--1490, 2010.

\bibitem[DJKV17]{storm}
Christian Dehnert, Sebastian Junges, Joost{-}Pieter Katoen, and Matthias Volk.
\newblock A storm is coming: {A} modern probabilistic model checker.
\newblock In Rupak Majumdar and Viktor Kuncak, editors, {\em Computer Aided
  Verification - 29th International Conference, {CAV} 2017, Heidelberg,
  Germany, July 24-28, 2017, Proceedings, Part {II}}, volume 10427 of {\em
  Lecture Notes in Computer Science}, pages 592--600. Springer, 2017.

\bibitem[DJL{\etalchar{+}}15]{David2015}
Alexandre David, Peter~Gj{\o}l Jensen, Kim~Guldstrand Larsen, Marius
  Mikucionis, and Jakob~Haahr Taankvist.
\newblock Uppaal stratego.
\newblock In Christel Baier and Cesare Tinelli, editors, {\em Tools and
  Algorithms for the Construction and Analysis of Systems - 21st International
  Conference, {TACAS} 2015, Held as Part of the European Joint Conferences on
  Theory and Practice of Software, {ETAPS} 2015, London, UK, April 11-18, 2015.
  Proceedings}, volume 9035 of {\em Lecture Notes in Computer Science}, pages
  206--211. Springer, 2015.

\bibitem[DS02]{DeCoste2002}
Dennis DeCoste and Bernhard Sch{\"{o}}lkopf.
\newblock Training invariant support vector machines.
\newblock {\em Mach. Learn.}, 46(1-3):161--190, 2002.

\bibitem[FCH{\etalchar{+}}08]{liblinear}
Rong{-}En Fan, Kai{-}Wei Chang, Cho{-}Jui Hsieh, Xiang{-}Rui Wang, and
  Chih{-}Jen Lin.
\newblock {LIBLINEAR:} {A} library for large linear classification.
\newblock {\em J. Mach. Learn. Res.}, 9:1871--1874, 2008.

\bibitem[HKP{\etalchar{+}}19]{Hartmanns2019}
Arnd Hartmanns, Michaela Klauck, David Parker, Tim Quatmann, and Enno Ruijters.
\newblock The quantitative verification benchmark set.
\newblock In Tom{\'{a}}s Vojnar and Lijun Zhang, editors, {\em Tools and
  Algorithms for the Construction and Analysis of Systems - 25th International
  Conference, {TACAS} 2019, Held as Part of the European Joint Conferences on
  Theory and Practice of Software, {ETAPS} 2019, Prague, Czech Republic, April
  6-11, 2019, Proceedings, Part {I}}, volume 11427 of {\em Lecture Notes in
  Computer Science}, pages 344--350. Springer, 2019.

\bibitem[HTF09]{Hastie2009}
Trevor Hastie, Robert Tibshirani, and Jerome~H. Friedman.
\newblock {\em The Elements of Statistical Learning: Data Mining, Inference,
  and Prediction, 2nd Edition}.
\newblock Springer Series in Statistics. Springer, 2009.

\bibitem[IS96]{Ittner1996}
Andreas Ittner and Michael Schlosser.
\newblock Non-linear decision trees - {NDT}.
\newblock In Lorenza Saitta, editor, {\em Machine Learning, Proceedings of the
  Thirteenth International Conference {(ICML} '96), Bari, Italy, July 3-6,
  1996}, pages 252--257. Morgan Kaufmann, 1996.

\bibitem[Jac20]{jackermeier}
M.~Jackermeier.
\newblock dtcontrol: Decision tree learning for explainable controller
  representation.
\newblock Bachelor's thesis, Technical University of Munich, 2020.

\bibitem[JAR{\etalchar{+}}20]{Jagtap2020}
Pushpak Jagtap, Fardin Abdi, Matthias Rungger, Majid Zamani, and Marco Caccamo.
\newblock Software fault tolerance for cyber-physical systems via full system
  restart.
\newblock {\em {ACM} Trans. Cyber Phys. Syst.}, 4(4):47:1--47:20, 2020.

\bibitem[JZ17]{quest}
Pushpak Jagtap and Majid Zamani.
\newblock {QUEST:} {A} tool for state-space quantization-free synthesis of
  symbolic controllers.
\newblock In Nathalie Bertrand and Luca Bortolussi, editors, {\em Quantitative
  Evaluation of Systems - 14th International Conference, {QEST} 2017, Berlin,
  Germany, September 5-7, 2017, Proceedings}, volume 10503 of {\em Lecture
  Notes in Computer Science}, pages 309--313. Springer, 2017.

\bibitem[Jü21]{thesisFiles}
Florian Jüngermann.
\newblock {Learning Algebraic Predicates for Explainable Controllers:
  Artifacts}.
\newblock \url{https://doi.org/10.5281/zenodo.4746131}, May 2021.

\bibitem[KNP11]{prism}
Marta~Z. Kwiatkowska, Gethin Norman, and David Parker.
\newblock {PRISM} 4.0: Verification of probabilistic real-time systems.
\newblock In Ganesh Gopalakrishnan and Shaz Qadeer, editors, {\em Computer
  Aided Verification - 23rd International Conference, {CAV} 2011, Snowbird, UT,
  USA, July 14-20, 2011. Proceedings}, volume 6806 of {\em Lecture Notes in
  Computer Science}, pages 585--591. Springer, 2011.

\bibitem[KZ19]{pfaces}
Mahmoud Khaled and Majid Zamani.
\newblock pfaces: an acceleration ecosystem for symbolic control.
\newblock In Necmiye Ozay and Pavithra Prabhakar, editors, {\em Proceedings of
  the 22nd {ACM} International Conference on Hybrid Systems: Computation and
  Control, {HSCC} 2019, Montreal, QC, Canada, April 16-18, 2019}, pages
  252--257. {ACM}, 2019.

\bibitem[Lev44]{Levenberg1944}
Kenneth Levenberg.
\newblock A method for the solution of certain non-linear problems in least
  squares.
\newblock {\em Quarterly of applied mathematics}, 2(2):164--168, 1944.

\bibitem[LMT15]{cruise}
Kim~Guldstrand Larsen, Marius Mikucionis, and Jakob~Haahr Taankvist.
\newblock Safe and optimal adaptive cruise control.
\newblock In Roland Meyer, Andr{\'{e}} Platzer, and Heike Wehrheim, editors,
  {\em Correct System Design - Symposium in Honor of Ernst-R{\"{u}}diger
  Olderog on the Occasion of His 60th Birthday, Oldenburg, Germany, September
  8-9, 2015. Proceedings}, volume 9360 of {\em Lecture Notes in Computer
  Science}, pages 260--277. Springer, 2015.

\bibitem[Mar63]{Marquardt1963}
Donald~W Marquardt.
\newblock An algorithm for least-squares estimation of nonlinear parameters.
\newblock {\em Journal of the society for Industrial and Applied Mathematics},
  11(2):431--441, 1963.

\bibitem[Mit97]{Mitchell1997}
Tom~M Mitchell.
\newblock {\em Machine learning}.
\newblock McGraw-hill New York, 1997.

\bibitem[MKS94]{Murthy1994}
Sreerama~K. Murthy, Simon Kasif, and Steven Salzberg.
\newblock A system for induction of oblique decision trees.
\newblock {\em J. Artif. Intell. Res.}, 2:1--32, 1994.

\bibitem[PVG{\etalchar{+}}11]{scikit-learn}
Fabian Pedregosa, Ga{\"{e}}l Varoquaux, Alexandre Gramfort, Vincent Michel,
  Bertrand Thirion, Olivier Grisel, Mathieu Blondel, Peter Prettenhofer, Ron
  Weiss, Vincent Dubourg, Jake VanderPlas, Alexandre Passos, David Cournapeau,
  Matthieu Brucher, Matthieu Perrot, and Edouard Duchesnay.
\newblock Scikit-learn: Machine learning in python.
\newblock {\em J. Mach. Learn. Res.}, 12:2825--2830, 2011.

\bibitem[PWH{\etalchar{+}}04]{Pradhan2004}
Sameer~S. Pradhan, Wayne~H. Ward, Kadri Hacioglu, James~H. Martin, and Daniel
  Jurafsky.
\newblock Shallow semantic parsing using support vector machines.
\newblock In Julia Hirschberg, Susan~T. Dumais, Daniel Marcu, and Salim Roukos,
  editors, {\em Human Language Technology Conference of the North American
  Chapter of the Association for Computational Linguistics, {HLT-NAACL} 2004,
  Boston, Massachusetts, USA, May 2-7, 2004}, pages 233--240. The Association
  for Computational Linguistics, 2004.

\bibitem[Qui86]{id3}
J.~Ross Quinlan.
\newblock Induction of decision trees.
\newblock {\em Mach. Learn.}, 1(1):81--106, 1986.

\bibitem[Qui93]{c45}
J.~Ross Quinlan.
\newblock {\em {C4.5:} Programs for Machine Learning}.
\newblock Morgan Kaufmann, 1993.

\bibitem[RWR15]{Rungger2015}
Matthias Rungger, Alexander Weber, and Gunther Reissig.
\newblock State space grids for low complexity abstractions.
\newblock In {\em 54th {IEEE} Conference on Decision and Control, {CDC} 2015,
  Osaka, Japan, December 15-18, 2015}, pages 6139--6146. {IEEE}, 2015.

\bibitem[RZ16]{scots}
Matthias Rungger and Majid Zamani.
\newblock {SCOTS:} {A} tool for the synthesis of symbolic controllers.
\newblock In Alessandro Abate and Georgios~E. Fainekos, editors, {\em
  Proceedings of the 19th International Conference on Hybrid Systems:
  Computation and Control, {HSCC} 2016, Vienna, Austria, April 12-14, 2016},
  pages 99--104. {ACM}, 2016.

\bibitem[SHB00]{APRICODD}
Robert St{-}Aubin, Jesse Hoey, and Craig Boutilier.
\newblock {APRICODD:} approximate policy construction using decision diagrams.
\newblock In Todd~K. Leen, Thomas~G. Dietterich, and Volker Tresp, editors,
  {\em Advances in Neural Information Processing Systems 13, Papers from Neural
  Information Processing Systems {(NIPS)} 2000, Denver, CO, {USA}}, pages
  1089--1095. {MIT} Press, 2000.

\bibitem[SZ19]{Swikir2019}
Abdalla Swikir and Majid Zamani.
\newblock Compositional synthesis of symbolic models for networks of switched
  systems.
\newblock {\em {IEEE} Control. Syst. Lett.}, 3(4):1056--1061, 2019.

\bibitem[Wei20]{weinhuber}
Christoph Weinhuber.
\newblock Learning domain-specific predicates in decision trees for explainable
  controller representation.
\newblock Bachelor's thesis, Technical University of Munich, 2020.

\bibitem[ZVJ18]{Zapreev2018}
Ivan~S. Zapreev, Cees Verdier, and Manuel~Mazo Jr.
\newblock Optimal symbolic controllers determinization for {BDD} storage.
\newblock In {\em {ADHS}}, volume 51-16 of {\em IFAC-PapersOnLine}, pages 1--6.
  Elsevier, 2018.

\end{thebibliography}

\clearpage
\arxivref{
\appendix

\section*{Appendix}

\section{Running Example}\label{app:runningExample}

Here we provide technical details on the cruise model.

\subsection{Cruise Control Modifications} \label{sec:cruiseModifications}
The cruise control system is modeled in the tool \t{UPPAAL Stratego} \cite{David2015}. The model was introduced in \cite{cruise} and is available to download on the respective website \url{https://people.cs.aau.dk/~marius/stratego/cruise.html}. We made the following small adjustment as in previous work \cite{dtControl,dtcontrol2}. As the accelerations are $-2, 0$ or $2$ and the time step is $1$, all velocities occurring in the model are even. However, when a car appears from the far-away state, it can also have an odd number velocity. To keep the even velocities, we changed the source code in lines 242ff. to:
\begin{lstlisting}[numbers=left, firstnumber=242, xleftmargin=.18\textwidth, xrightmargin=.18\textwidth]
	i:int[minVelocityFront/2, maxVelocityFront/2]
	2*i &lt;= velocityEgo
	velocityFront = i*2,
	distance = maxSensorDistance,
	rVelocityFront = 2 * i * 1.0,
	rDistance = 1.0*maxSensorDistance
\end{lstlisting}
The modified model file can be found in \cite{thesisFiles}.

\subsection{Cruise Control Parameters} \label{sec:cruiseParameters}
We made some further adjustments to the \cruise\ model compared to the version available at \url{https://people.cs.aau.dk/~marius/stratego/cruise.html} and different from previous work \cite{dtControl,dtcontrol2}. In addition to the modifications previous works made (described in Appendix \ref{sec:cruiseModifications}) we also changed the constants $v_{min}, v_{max},$ and $d_{max}$. The \cruise\ model used in \cite{dtControl} and \cite{dtcontrol2} specified $v_{max} = 20$, $v_{min} = -10$, and $d_{max} = 200$. This provoked the following unwanted behavior: When the distance between the the front and the ego vehicle is at around $150$, both cars can drive at full speed. However, when the relative distance approaches $200$, the ego vehicle needs to slow down. The reason is the following behavior of the front vehicle:
\begin{enumerate}
	\item the front vehicle drives with $v_f = 20$ and $d_r=190$,
	\item the front vehicle disappears into the far-away state as $d_r > 200$,
	\item a \enquote{new} car appears at the end of the sensor range $d_r=200$. Independent of the velocity the front vehicle had before, the new car can have any velocity, for example $v_f = -10$.
\end{enumerate}
This way, the front vehicle effectively changes its velocity from $v_f=20$ to $v_f=-10$ in just a couple of time steps. Even the distance of $200$ is not enough for the ego vehicle to react and avoid a crash in this scenario. We fix this flaw by increasing the minimal velocity and the maximum sensor distance so that the ego vehicle has enough time to break if a new car suddenly appears. 

\autoref{tab:cruise_parameters} contains an overview of the parameters used and the resulting size of the controller measured in number of states and number of state-action pairs. The generated controllers are included in \cite{thesisFiles}.
\begin{table}[t]
	\centering
	\caption{The parameters used for generating the controllers of the cruise model and the resulting sizes measured in number of states and number of state-action pairs.}
	\begin{tabular}{@{}lrrrrr@{}} \toprule
		& \multicolumn{3}{c}{Parameters} & \multicolumn{2}{c}{Controller Size} \\
		\cmidrule(r){2-4} \cmidrule(r){5-6}
		Name & $v_{min}$ & $v_{max}$ & $d_{max}$ & \#states & \#s-a pairs\\\midrule
		\t{cruise\_prev}& -10 & 20 & 200 & 295,615 & 886,845\\
		\t{cruise\_250} & -6 & 20 & 250 & 320,523 & 961,569\\
		\t{cruise\_300} & -10 & 20 & 300 & 500,920 & 1,502,760\\ \bottomrule
	\end{tabular}
	\label{tab:cruise_parameters}
\end{table}

\subsection{Handwritten Controller}\label{app:hand}

To better understand the model, we will briefly explain how the handcrafted strategy works. In the worst case, the front vehicle will start decelerating in the next time step and will continue until it has reached its minimal velocity. For our car, we have to decide what action to take for the next time step $t_1$: accelerate, stay neutral or decelerate. To see if it is safe to accelerate, we calculate the relative distance after accelerating for one time step $t_1$ and then decelerating until the ego vehicle has reached the minimal velocity.

In \autoref{fig:handcrafted_geometric} we have plotted the velocity-time diagram describing the kinematics of both cars in case the ego vehicle accelerates in the next time step. The front vehicle (red) instantly decelerates with the rate $a_{min}$ and then continues with minimal velocity. The ego vehicle (blue) starts with a higher velocity, accelerates for one step, and then decelerates with the same rate. The distance traveled is the time integral of the velocity, so the area between the curves describes the relative distance change. We can partition the area into four sections and calculate the respective areas:
\begin{align*}
	A_1 &= -\frac{v_e^2}{2a_{min}} - \left(- \frac{v_f^2}{2a_{min}} \right) \\
	A_2 &= v_{min}\frac{v_e - v_f}{a_{min}}\\
	A_3 &= a_{max} t_1^2 \left(1- \frac{a_{max}}{a_{min}}\right)\\
	A_4 &= \left(v_e - v_{min} \right) t_1 \left(1- \frac{a_{max}}{a_{min}}\right)
\end{align*} 
With these values, we can write the predicate deciding whether it is safe to accelerate in the next time step as a quadratic polynomial of our state variables $v_e, v_f, d_r$. Note that \autoref{eq:handcraftedPoly2} is the same as \autoref{eq:handcraftedPoly}.
	\begin{align} \label{eq:handcraftedPoly2}
		\frac{1}{2a_{min}} v_e^2 
		&- \frac{1}{2a_{min}} v_f^2 
		- \left(\frac{v_{min}}{a_{min}} + t_1(1 - \frac{a_{max}}{a_{min}})\right) v_e 
		+ \frac{v_{min}}{a_{min}} v_f 
		-\notag \\
		& \left(1 - \frac{a_{max}}{a_{min}}\right) t_1(t_1 a_{max} - v_{min}) 
		+\, d_r
		~~~~\ge ~~~~ d_{safe}
	\end{align}

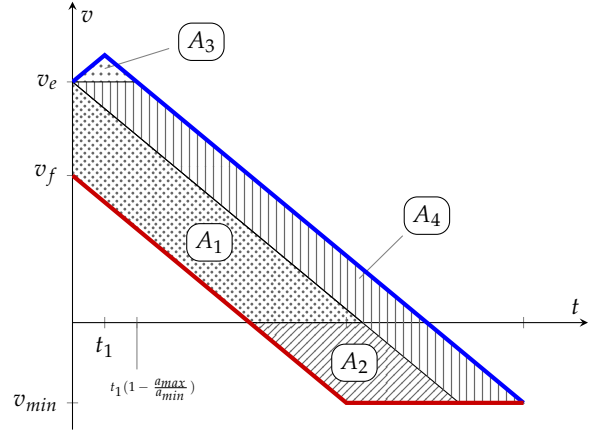
\begin{figure}[t]
	\centering
	\definecolor{myred}{HTML}{c80000}
\definecolor{mygreen}{HTML}{33652b}
\definecolor{myorange}{HTML}{c07400}
\tikzstyle{v} = [ultra thick]
\tikzstyle{vF} = [v, myred]
\tikzstyle{vE} = [v, blue]
\tikzstyle{A} = [draw, fill=white, rounded corners=5pt]

\begin{tikzpicture}
  \begin{axis}[
    axis lines=center,
    ytick={-6, 11, 18},
    yticklabels={$v_{min}$, $v_f$, $v_e$},
    xtick={2, 4, 17, 28},
    xticklabels={$t_1$, , , ,}, 
    xmin=0,
    xmax=32,
    ymax=24,
    ymin=-8,
    xlabel=$t$,
    ylabel=$v$,
  ]
    \draw[pattern=crosshatch dots, pattern color=black!60]
      (0, 11) -- (0, 18) -- (18, 0) -- (11, 0)
    ;
    \node[A] at (8.5, 6) {\small $A_1$};

    \draw[pattern=north east lines, pattern color=black!60]
      (18, 0) -- (24, -6) -- (17, -6) -- (11, 0)
    ;
    \node[A] at (17.5, -3) {\small $A_2$};
    
    \draw[pattern=dots, pattern color=black!60]
    (0, 18) -- (2, 20) -- (4, 18)
    ;
    \draw[gray] (2, 19) -- (6, 20.5);
    \node[A] at (8, 21) {\small $A_3$};

    \draw[pattern=vertical lines, pattern color=black!60]
      (0, 18) -- (4, 18) -- (28, -6) -- (24, -6)
    ;
    \draw[gray] (17.8, 3.3) -- (20.5, 6.5);
    \node[A] at (22, 8) {\small $A_4$};

    \draw (0, 18)  -- (18, 0);
    \draw[vE] (0, 18)  -- (2, 20) -- (2, 20)  -- (28, -6);
    
    \draw[vF] (0, 11)  -- (17, -6) -- (17, -6) -- (28, -6);
    
    \draw[gray] (4, 0) -- (4, -3.5);
    \node at (5, -5) {{\tiny$t_1(1-\frac{a_{max}}{a_{min}})$}};

  \end{axis}
\end{tikzpicture}
	\caption{A velocity-time graph showing the ego vehicle accelerating for one time step. The area between the blue and the red curve describes the change in distance between the two cars.}
	\label{fig:handcrafted_geometric}
\end{figure}

\section {Predicates From Domain Knowledge}

\subsection{Base Identities} \label{sec:allBaseIdentities}
These 8 base identities are used when generating new predicates:
\begin{align*}
	a &= \frac{2(d - tv)}{t^2} \\
	a &= \frac{v}{t} \\
	t &= \frac{-v + \sqrt{2ad + v^2}}{a} \\
	t &= -\frac{v + \sqrt{2ad + v^2}}{a} \\
	t &= \frac{v}{a} \\
	v &= -\frac{at}{2} + \frac{d}{t} \\
	v &= at \\
	d &= \frac{at^2}{2} + vt
\end{align*}

\subsection{Handcrafted Predicate Derivation} \label{app:handcraftedPredicateIntroduction}
In this section, we describe how to derive the predicate in \autoref{eq:handcraftedPoly} (described in more detail in Appendix~\ref{app:hand}) using the automated approach of Section~\ref{chapter:pred_domain_knowledge}. We utilize the following expressions:
\begin{itemize}
	\item $d_{one}$: the change in relative distance during the first time step.
	\item $v_{fChg}, v_{eChg}$: the change in velocity for both cars during the first time step.
	\item $v_{f}', v_{e}'$: the new velocity of both cars after the first time step.
	\item $t_{f}, t_{e}$: the time after the first time step until the front car reaches minimal velocity.
	\item $d_{f}, d_{e}$: the distance travelled by the respective cars after the first time step until the front car reaches minimal velocity.
	\item $d_{fe}$: the distance the front car travels with minimal velocity until the ego car also reaches its minimal velocity.
\end{itemize}
\autoref{fig:handcraftedPredicate} shows how we can arrive at the handcrafted predicate. At the top, we have a subset of the base values $S \cup C$ comprising constants like $a_{max}$ and state variable like $d_r$. The color encodes to which physical quantity a value belongs. For example, all velocities are drawn in blue. Then the diagram shows the iterations of our algorithm. Every iteration consists of two phases: Step 3 where values of the same type can be added or subtracted to form new values, and Step 4 where we use our domain knowledge base identities to calculate new values. We leave out the irrelevant values as the total number of generated values would be far too large as we will see in the next section.

We observe that the first meaningful distance predicates emerge after four iterations. Then, summing the distance expressions together takes another 3 iterations. The non-simplified version of the predicate that our algorithm would output is 
\begin{align*}
	&d_{one} + d_{f} - d_{e} + d_{fe} + d_r \ge d_{safe}\\
	\Leftrightarrow \quad & \\
	&(a_{min}-a_{max})t_1^2/2 + (v_f-v_e)t_1 \\
	+ &a_{min}((v_{min}-(v_f + a_{min}t_1))/a_{min})^2/2 + (v_f + a_{min}t_1)(v_{min}-(v_f + a_{min}t_1))/a_{min} \\
	- &a_{min}((v_{min}-(v_e + a_{max}t_1))/a_{min})^2/2 - (v_e + a_{max}t_1)(v_{min}-(v_e + a_{max}t_1))/a_{min} \\
	+ &v_{min}  \left[(v_{min}-(v_e + a_{max}t_1))/a_{min} - (v_{min}-(v_f + a_{min}t_1))/a_{min}\right] \\ 
	+ &d_r \ge d_{safe}
\end{align*}

By reformulating, we see that this is the same predicate as given in \autoref{eq:handcraftedPoly} and derived in Appendix~\ref{app:hand}.
There is only a slight difference in the constant offset.
The easiest way to validate this predicate is by using the understanding given in \autoref{fig:handcrafted_geometric}. For now, we assume $v_e \ge v_f$ and we accelerate while the front car brakes, as this is the worst case. 
Then, the kinematics can be described in three phases. First, the behavior in the next time step where the ego vehicle accelerates and the front vehicle decelerates ($d_{one}$). Second, the phase when both cars break lasting until the front car has reached minimum velocity ($d_{f1}-d_e$). Third, the final phase where the front car continues at minimal velocity whereas the ego car continues to decelerate until it also reaches minimum velocity $d_{fe}$.

\subsection{Identified Problems} \label{app:identified_problems}
We have identified two fundamental problems with this approach: the large search space and the missing uniqueness of the derivation.

First, as the search space is so large, we need a heuristic telling us which expression will be useful at a later stage. When introducing the approach we stressed that we can use any intermediate predicate in the decision tree. So a natural choice for a heuristic would be some kind of impurity measure on the controller data. Unfortunately, expressions like the time until we reach minimal velocity are not great splitting predicates. And even expressions close to the final predicate like $d_{one}$ or $d_{e}$ are of little use on their own. In \autoref{fig:predicate_2d}, we plot the handcrafted predicate $d^*$ together with its individual components in a two-dimensional plot for the value $v_f = 2$. While each predicate individually is a bad classifier, their sum can perfectly classify the data. Given only the impurities of the predicates, there does not seem to be a way to conclude which predicates are useful. Thus, developing a better heuristic is an important step for future work.

\begin{figure}[ht]
	\centering
	\input{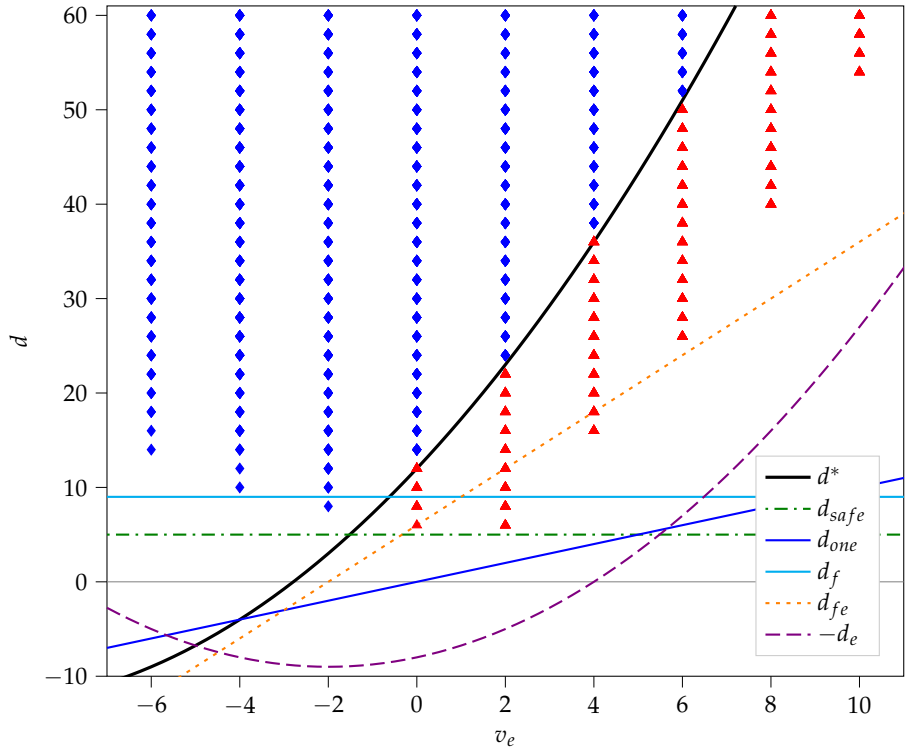}
	\caption{The handcrafted predicate $d^*$ and the terms used to derive it, plotted for a fixed value of $v_f = 2$. While the sum of the terms is a perfect classifier, individually, they are not helpful for the classification.}
	\label{fig:predicate_2d}
\end{figure}

Second, the handcrafted predicate is so complex that there are alternative expressions that evaluate to the same predicate after substituting the constants with their values. For example, when calculating $d_{one}$, 
we set the acceleration to $a_{min} - a_{max}$ which evaluates to $-4$. Our approach would also try setting the acceleration to $a_{min} + a_{min}$ which also evaluates to $-4$ but lacks any relation to the situation we want to describe. So only the general domain knowledge and the controller data might not be enough to unambiguously derive an explainable predicate. Either a human has to steer the process and select the most sensible predicates, or we have to somehow incorporate additional information about the model.

\begin{figure}[ht]
	\centering
	\subfloat[][]{ \label{fig:curve_fitting_2d}
		\begin{tikzpicture}
  \begin{axis}[
    width=0.5\textwidth ,
	  scatter/classes={
      a={mark=square*,blue},
      b={mark=*,red}
    }
  ]
    \addplot[
      scatter,
      only marks,
      scatter src=explicit symbolic
    ]
    table[meta=label]
    {
      x   y   label
      1   1   a 
      1   2   a
      1   3   a
      1   4   a
      2   1   b
      2   2   b
      2   3   a
      2   4   a
      3   1   b
      3   2   b
      3   3   b
      3   4   a
      4   1   b
      4   2   b
      4   3   b
      4   4   b
    };
    \addplot [
      domain=0.5:4.3,
      restrict y to domain=0.5:4.5,
      color=black,
      thick,
    ] {-0.3*(x-4.6)^2 + 4.3};
  \end{axis}
\end{tikzpicture}
	}
	\subfloat[][]{ \label{fig:curve_fitting_3d}
		\begin{tikzpicture}
  \begin{axis}[
    width=0.5\textwidth ,
    view={134}{10},
    xmin=0.5, xmax=4.5,
    ymin=0.5, ymax=4.5,
    zmin=-2, zmax=2,
	  scatter/classes={
      a={mark=square*,blue},
      ah={mark=square*, blue, fill opacity=0.2, draw opacity = 0},
      b={mark=*,red},
      bh={mark=*,red, fill opacity=0.3, draw opacity = 0}
    }
  ]
    \addplot3[
      scatter,
      only marks,
      scatter src=explicit symbolic
    ]
    table[meta=label]
    {
      x   y   z label
      1   1   1 ah
      1   2   1 a
      1   3   1 a
      1   4   1 a
      2   1  -1 b
      2   2  -1 bh
      2   3   1 a
      2   4   1 a
      3   1  -1 b
      3   2  -1 b
      3   3  -1 bh
      3   4   1 a
      4   1  -1 b
      4   2  -1 b
      4   3  -1 b
      4   4  -1 bh
    };
    \addplot3[
      mesh,
      domain=0.5:4.5,
      colormap/cool, 
      colormap={bluegray}{rgb255=(100,100,150) rgb255=(170,220,255)},
    ] {-tanh(2*(-0.3*(x-4.6)^2 + 4.3 - y))};
    
    \addplot3[
      surf,
      shader=flat,
      draw=black!60,
      fill=gray,
      opacity=0.6,
      fill opacity=0.4,
      domain=0.5:4.5,
      domain y =0.5:4.5,
      samples=80,
      samples y =80,
      restrict z to domain=-2: 2,
    ] {-3*tanh(2*(-0.3*(x-4.6)^2 + 4.3 - y))};
  \end{axis}
\end{tikzpicture} 
	}
	\caption{In the current curve-fitting implementation, a two-dimensional dataset (a) is mapped to a three-dimensional space where the $z \in \{-1, 1 \}$ is determined by the label. The old approach then fits a function to the new dataset. Our approach instead tries to separate the data like the gray surface does in (b).}
	\label{fig:curve_fitting}
\end{figure}
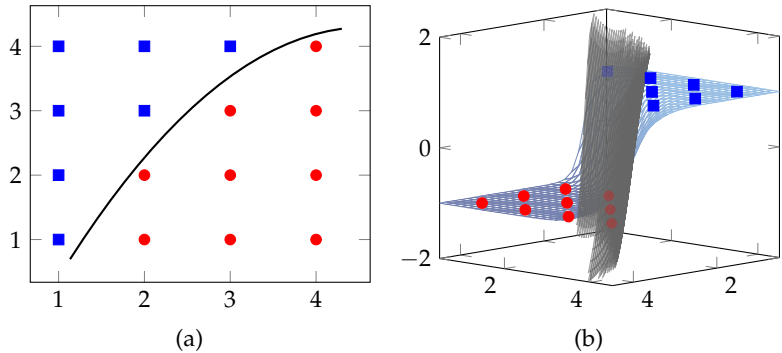

\section{Predicates From Controller Data}

By looking at a two-dimensional plot for a fixed value of $v_f$ in \autoref{fig:predicate_2d}, we see that the predicate is well defined by the data points -- we should be able to reconstruct the splitting function by solely looking at the data and fitting a function to it.

\subsection{Problems with Curve Fitting}\label{sec:problem_with_curve_fitting}
The recent extensions of \dtControl\ \cite{weinhuber,dtcontrol2} enable us to use curve fitting \parencite{curvefittingBook} for finding unspecified coefficients. We know from \autoref{eq:handcraftedPoly} that the handpicked strategy is a quadratic polynomial so we can try to use a general quadratic polynomial $c_1 v_e^2 + c_2 v_e v_f + c_3 v_f^2 + \ldots$ and determine the coefficients with curve fitting. Unfortunately, this approach fails to find the correct predicate. To understand why we need to investigate how the curve fitting is implemented.

For now, we always consider a \i{one versus the rest} split. This means, we pick a label $y$ that we want to separate from the rest and set
$$ y_i' = 
\begin{cases}
	+1  & \text{if } y_i = y \\
	-1 & \text{else}
\end{cases}
\quad
\text{for all } i
$$

Consider the two-dimensional data from \autoref{fig:curve_fitting_2d}. What the current version of curve fitting does is the following. First, we map our two-dimensional data $x_i \in \mathbb{R}^2$ with label $y_i' \in \{-1, 1\}$ to the three-dimensional space where $y_i'$ is used as the third coordinate. Then we use regression analysis to fit a function to the data with least-squares-fitting \parencite{Levenberg1944,Marquardt1963} (see \autoref{fig:curve_fitting_3d}). What we propose in this thesis, is to use a classification approach rather than a regression approach. So in \autoref{fig:curve_fitting_3d} we are interested in the gray function \i{separating} the data points instead of fitting them. This way, we put most emphasis on the sample points close to the split rather than weighting every sample equally. Coming back to the two-dimensional space (\autoref{fig:curve_fitting_2d}), we want to find a function that smoothly separates the labels, ideally maximizing the distance to any specific sample. This is where support vector machines come into play.

\subsection{Using Support Vector Machines}\label{app:svm}
Support vector machines (SVMs) exactly do what we want here: find a function that separates the data and maximizes the margins. The main idea offered in this work is that we can reconstruct the algebraic decision function from the internal coefficients of the SVM. This is not feasible for tools like neural networks \cite[Chapter 11]{Hastie2009} but we will see how and under what conditions it is possible for SVMs in the next sections.

The \dtControl\ tool already supports finding linear splitting predicates with SVMs. However, for the \cruise\ example, a linear predicate is not enough to perfectly split the data. So we are tempted to use a polynomial kernel to increase the expressiveness of our SVM. However, 
the runtime of common training algorithms for SVMs is at least quadratic in the number of samples. And in fact, the algorithms implemented in the open-source tool \t{scikit-learn} \cite{scikit-learn} do not terminate within an hour for the \cruise\ example with a few hundred thousand sample points.

We can circumvent this issue by taking advantage of our specific use case. Usually, SVMs are used with high-dimensional datasets like images \cite{DeCoste2002} or language models \cite{Pradhan2004} where the number of features has the same order of magnitude as the number of samples. For the purpose of controller synthesis, the number of state variables is usually small as the number of states usually grows exponentially with the number of state variables. So while the kernel trick is useful for high-dimensional data, we can renounce the kernel trick in our case and explicitly construct the higher-dimensional space. 
A similar idea is also described in \cite{Chang2010}.

For example, when we want to change from the linear three-dimensional space $(v_e, v_f, d_r)^T$ to the quadratic space, we will have the following 9 dimensions
\begin{align} 
	(v_e,\, v_f,\, d_r,\, v_e v_f,\, v_e d_r,\, v_f d_r,\, v_e^2,\, v_f^2,\, d_r^2)^T
\end{align}
The moderate increase in dimensions is clearly outweighed by the much better performance of the linear SVM algorithms we can now use.

\subsubsection{Problems With Higher Dimensions} \label{sec:problemsWithHigherDimensions}
At the moment, we only support mapping to the quadratic space, which means our predicates are quadratic polynomials. For higher degree polynomials, we have not seen that the gained expressiveness justifies the significantly increased complexity of the predicates. For example, a cubic predicate with $5$ variables already has $55$ terms. Even with the methods we will discuss in \autoref{sec:featureImportance} and \autoref{sec:roundingCoefficients}, this predicate will not fulfil our goal of being explainable. Mapping to a space with features like $e^x$ or $\sin(x)$ poses the challenge that we can only fit the coefficient, but not scale the function in $x$-direction like $e^{cx}$ or $\sin (cx)$ and is therefore left for future work.

\subsubsection{Reconstructing the Algebraic Decision Function}
Assuming that our SVM finds a separating hyperplane that we want to use as a splitting predicate in our decision tree, how do we reconstruct the algebraic representation? The SVM algorithm finds a hyperplane $(\vec{w}^*, b^*)$ with $\vec{w}^*\cdot \vec{x} - b^* = 0$ where $\vec{x}$ corresponds to a transformed set of state variables in the form of \autoref{eq:highDimSpace}. This means the $w_i$ are the coefficients of the quadratic polynomial of our state variables.

When implementing it in practice, there is a small intermediate step we want to mention for completeness. In order for the quadratic optimization algorithm to work properly, the input data needs to be normalized to have a mean of 0 and a standard deviation of 1. This standardization of course has to be taken into account when exporting the coefficients.



\subsection{Predicate Without Prettifying} \label{sec:prePrettifyEq}
Before applying the methods described in \autoref{sec:featureImportance} and \autoref{sec:roundingCoefficients}, a predicate for the \cruise\ example looks like this (rounded to 6 decimal places):
\begin{align*}
	&-1.004058e_{choose}d_r+0.000121d_r^2
	+4.011296e_{choose}v_e
	-0.002316d_rv_e
	+0.51353v_e^2\\
	&+8.5\cdot 10^{-5}e_{choose}a_e
	-0.000276d_ra_e
	+0.002239v_ea_e
	-6.4\cdot 10^{-5}a_e^2\\
	&-3.007296e_{choose}v_f
	+0.001317d_rv_f
	-0.012358v_ev_f
	-0.001334a_ev_f\\
	&-0.499783v_f^2
	-0.000224e_{choose}a_f
	+0.000261d_ra_f
	-0.002111v_ea_f\\
	&-6.4\cdot 10^{-5}a_ea_f
	+0.001224v_fa_f
	+3.1\cdot 10^{-5}a_f^2
	-1.004058d_r\\
	&+4.011296v_e
	+8.5\cdot 10^{-5}a_e
	-3.007296v_f
	-0.000224a_f\\
	&+23.107387 \le 0
\end{align*}

\subsection{Predicate Without Rounding} \label{sec:preRounding}
After leaving out unimportant features as described in \autoref{sec:featureImportance} but without rounding coefficients  as described in \autoref{sec:roundingCoefficients}, a predicate for the \cruise\ example looks like this (rounded to 6 decimal places):
\begin{align*}
	&-0.000463d_r^2
	+0.008656d_rv_e
	-0.549255v_e^2
	-0.005078d_rv_f\\
	&+0.046916v_ev_f
	+0.496888v_f^2
	+2.043519d_r
	-10.25286v_e\\
	&+6.138132v_f
	-39.685041 \le 0
\end{align*}


\subsection{Numerical Errors}\label{app:num}
One problem we need to handle concerns floating point precision errors. When testing our classifier, we use the internal coefficients of the SVM. The coefficients we output are different though, as we need to undo the normalization we applied. We must ensure that possible precision errors from these transformations do not change the classification.
In the original predicate generated by the SVM, the classifier maximizes the margin between the label sets so we can be quite confident that small precision errors will not change the classification\footnote{Note that this only holds for cases where we find a perfect split.}. When trying out rounded coefficients, however, we lose this property. A rounded coefficient might classify everything correctly but the slightly different transformed coefficient might lead to other results.

As a heuristic against these problems, we over-approximate a change when trying out a rounded coefficient. For example, if our current coefficient is $2.953$ and we want to try the rounded value $3$, we over-approximate the change and try $3.00001$ instead. If that works, we can be more confident that the value $3$ will not lead to those problems.

Additionally, we also encounter precision problems if our SVM uses very large coefficients. How we deal with them is discussed in Appendix \ref{sec:advancedNumProps}.

When the tree construction is finished, \dtControl\ verifies that every sample is classified correctly or outputs an error rate. In this step, we use the transformed coefficients of the polynomial we output so we can be sure that the decision tree is as accurate as the tool tells the user.

\subsection{Advanced Numerical Precision Problems} \label{sec:advancedNumProps}
Sometimes, when training the SVM, very large coefficients occur. As long as every datapoint is located on the right side of the hyperplane, the loss function 
does not penalize these large coefficients. However, when evaluating a predicate like $$2x_1 - 3 x_2 + 10^{18} x_3 - 10^{18} \le 0$$ the floating point precision reaches its limits. So, in the rare case that we observe such a behavior, we apply the following countermeasure.

For every feature $i$, we add the \i{control samples} $(\vec{e}_i, 1)$ and $(\vec{e}_i, -1)$ to the dataset, where $\vec{e}_i$ is the unit vector in direction of feature $i$. Then we re-train the SVM. Note that this way, the data is not linearly separable. The loss function 
penalizes control samples that are far away from the separating hyperplane because either the positive or the negative control sample is located on the wrong side. Hence, the SVM uses small coefficients to keep the control samples reasonably close to the decision function. The \t{liblinear} tool \cite{liblinear} that we use in this work also supports different sample weights. Thus, we give the control samples a weight that is three orders of magnitude smaller than the weights for the regular samples so we do not disturb the regular training too much.

If, for some reason, we still have coefficients with an absolute value larger than $10^7$ or smaller than $10^{-7}$ but not $0$, we change them to a value inside of this interval. This might change the classification but therefore ensures that we do not run into precision errors after exporting the decision tree and evaluating the predicate on a device with a slightly different floating-point engine.

\section{Further Experimental Results}

\subsection{Domain Knowledge Approach}\label{app:exp-domain}
As we have seen in \autoref{chapter:pred_domain_knowledge}, our approach was unable to generate the handcrafted predicate for the \cruise\ example. Still, we generated a lot of predicates that might be useful when building the decision tree. We evaluate three sets of predicates that we generated with our approach from \autoref{chapter:pred_domain_knowledge}. As the number of predicates increases so fast and we cannot even complete two iterations, we also try skipping Step 3 in the approach, meaning we do not add sums and differences of our values. The number of nodes of the resulting decision trees for the \t{cruise\_250} dataset are shown in \autoref{tab:results_domain_knowledge}. In addition to the number of generated predicates, we also list the number of unique predicates as multiple predicates can evaluate to the same expression after substituting the constants with their values (as we have seen in the example with $a_{min} = -a_{max}$). 
To build the decision tree, we use the generated predicates in addition to the axis-aligned predicates and choose the best splitting predicates using the entropy impurity measure.

As a comparison, we have included the decision trees we receive without domain knowledge using only axis-aligned splits and using linear predicates generated with the OC1 heuristic \cite{Murthy1994}.

\begin{table}
	\centering
	\small
	\caption{The decision tree sizes (number of nodes) while using different sets of predicates for the \t{cruise\_250} dataset. The column \enquote{Sum?} describes if we use Step 3 of our generation approach from \autoref{chapter:pred_domain_knowledge}.}
	\begin{tabular}{@{}rr rr r@{}} \toprule
  Iterations & Sum? & \#Predicates & \#Unique Pred. & DT Size\\
  \midrule
           1 &   No &          66 &            42 & 655 \\
           1 &  Yes &       3,604 &         1,929 & 395 \\
           2 &   No &      10,568 &         9,634 & 269 \\
  \multicolumn{5}{l}{\textbf{Comparision}} \\
  \multicolumn{4}{l}{Axis-aligned predicates}     & 869\\
  \multicolumn{4}{l}{Linear predicates}           & 369\\
  \bottomrule
\end{tabular}
	\label{tab:results_domain_knowledge}
\end{table}

\medskip
We see that the generated predicates help find succinct decision trees. For the largest predicate set, we even find a smaller tree than we do with linear predicates. Still, it is not clear whether this is because the predicates describe the dynamics of the system well or whether this improvement is simply due to the large number of predicates we try. In fact, we try so many splitting predicates that the runtime increases from 1 minute when using linear predicates to over twelve hours for the large predicate set, even after implementing use-case-specific optimizations.

As we did not succeed in creating truly explainable decision trees for the cruise example, we do not try this approach on other case studies but instead focussed on the second approach.

\subsection{Minimum Tree Size}\label{app:min}
To better understand the quality of our results, we compare them to the theoretical minimum-sized decision trees.
We can give a lower bound on the number of nodes the decision tree must contain if we want to represent the entire controller without determinizing (i.e.\ the most-permissive controller) as follows: At every state $s$, a subset of actions $C(s) \subseteq \mathcal{A}$ is allowed. We define $U \mathrel{:=} \{C(s)\, |\, s \in \mathcal{S}\}$ as the set of all possible allowed action subsets that occur in our controller at at least one state. To completely represent the controller, we need at least one distinct leaf in our decision tree for every distinct element $u \in U$. If we disregard the non-binary splits for categorical variables introduced in \cite{dtcontrol2}, we always build a full binary decision tree. As a full binary tree with $n$ leaves has $n-1$ inner nodes, the lower bound for the total number of nodes of our decision tree is $2|U|-1$.

If the decision predicates are sufficiently complex we can always achieve this bound. However, in practice this is often not even desirable. For example, we see that our decision tree for the cruise example in \autoref{fig:dt_cruise_250} does not have minimum size as it contains two leaves with the actions $a\in \{-2, 0\}$. Still, to keep an explainable decision tree, we would not want to merge those leaves as one describes that the car cannot accelerate because it has already reached its maximum velocity while in the other case, accelerating would be technically possible but would lead to unsafe behavior.

\subsection{Detailed Experimental Results} \label{chp:all_results}
Here we list the benchmark results for all case studies. The structure is the same as described in \autoref{sec:benchmarks} with two slight differences. First, we also give the size of the binary decision diagram (BDD) representation. Second, we explicitly list the tree sizes we get with the linear support vector machine, the logistic regression, and the OC1 heuristic splitting strategies. In \autoref{tab:results_condensed} in the main body, we only show the minimum across those.

The BDD sizes for the cyber-physical system cases is the minimum number of nodes from 10 tries. For the quantitative verification case studies, we show the BDD sizes from \cite{dtcontrol2} which correspond to the minimum across 20 tries.

\medskip
The benchmarks are split into three tables. \autoref{tab:results_cps_all} contains the cyber-physical system case studies, \autoref{tab:results_storm_all1} and \autoref{tab:results_storm_all2} contain case studies from the quantitative verification benchmark set \cite{Hartmanns2019}.
\begin{landscape}
	\begin{table}
		\centering
		\small
		\caption{Benchmark results for the cyber-physical system case studies.}
		\begin{tabular}{@{}lrrr rrrrrr@{}} \toprule
  & \multicolumn{3}{c}{Comparision}
  &  & \multicolumn{3}{c}{Linear} & \multicolumn{2}{c}{Quadratic}\\
 \cmidrule(r){2-4} \cmidrule(r){5-5} \cmidrule(r){6-8} \cmidrule(r){9-10}
  Case Study & States & BDD & MinSize
   & Ax.Al. & LinSVM & LogReg & OC1   & Poly & PolyPrio1\\ \midrule
   
   \multirow{2}{*}[0pt]{cartpole \cite{Jagtap2020}}
   & \multirow{2}{*}[0pt]{271} & \multirow{2}{*}[0pt]{312} & \multirow{2}{*}[0pt]{\b{169}}
   & 253    & 247    & 199    & 183   & 243  & 189 \\
   &&&& 263    & 263    & 187    & 261   & \b{169}  & \b{169} \\\addlinespace[0.5em]
   
   \multirow{2}{*}[0pt]{10rooms \cite{quest}}
   & \multirow{2}{*}[0pt]{26,244} & \multirow{2}{*}[0pt]{168} & \multirow{2}{*}[0pt]{\b{49}}
   & 17,297 & 157    & 147    & 4,515 & 61   & 61 \\
   &&&& 17,297 & 121    & 107    & 7,455 & \b{49}   & \b{49} \\\addlinespace[0.5em]
   
   \multirow{2}{*}[0pt]{helicopter \cite{Jagtap2020}}
   & \multirow{2}{*}[0pt]{280,539} & \multirow{2}{*}[0pt]{\b{1,348}} & \multirow{2}{*}[0pt]{475}
   & 6,339  & 5,787  & 3,769  & TO   & 5,035& 3,787 \\
   &&&&9,649& 9,763  & 4,637  & TO   & TO  & TO \\\addlinespace[0.5em]
   
   \multirow{2}{*}[0pt]{cruise\_250 \cite{cruise}} 
    & \multirow{2}{*}[0pt]{320,523} & \multirow{2}{*}[0pt]{1,820} & \multirow{2}{*}[0pt]{9}
      & 869    & 721    & 557    & 369   & 353  & 37 \\
   &&&& 1,067  & 817    & 657    & 363   & \b{11}   & 25 \\\addlinespace[0.5em]
   
   \multirow{2}{*}[0pt]{cruise\_300 \cite{cruise}}
    & \multirow{2}{*}[0pt]{500,920} & \multirow{2}{*}[0pt]{2,229} & \multirow{2}{*}[0pt]{9}
      & 1,157  & 991    & 691    & 467   & 521  & 59 \\
   &&&& 1,343  & 1,035  & 881    & 507   & \b{13}   & 23 \\\addlinespace[0.5em]
   
   \multirow{2}{*}[0pt]{dcdc \cite{scots}}
    & \multirow{2}{*}[0pt]{593,089} & \multirow{2}{*}[0pt]{575} & \multirow{2}{*}[0pt]{5}
       & 271    & 279    & 139    & 179   & \b{129}  & 199 \\
    &&&& 265    & 265    & 173    & 179   & 147  & 273 \\\addlinespace[0.5em]
   
   \multirow{2}{*}[0pt]{truck\_trailer \cite{pfaces}} 
    & \multirow{2}{*}[0pt]{1,386,211} & \multirow{2}{*}[0pt]{\b{36,169}} & \multirow{2}{*}[0pt]{1,839}
       & 338,283& TO    & TO    & TO   & TO  & TO \\
    &&&& 366,411& TO    & TO    & TO   & TO  & TO \\
   
   \multirow{2}{*}[0pt]{aircraft \cite{Rungger2015}}
    & \multirow{2}{*}[0pt]{2,135,056} & \multirow{2}{*}[0pt]{\b{177,332}} & \multirow{2}{*}[0pt]{31}
       & 915,877& 916,685& TO    & TO   & 725,011& 602,335 \\
    &&&&1,015,903&1,013,949& TO  & TO   & 688,577& 630,631 \\\addlinespace[0.5em]
   
   \multirow{2}{*}[0pt]{traffic\_30m \cite{Swikir2019}}
    & \multirow{2}{*}[0pt]{16,639,662} & \multirow{2}{*}[0pt]{TO} & \multirow{2}{*}[0pt]{23}
       & 12,573 & 9,631  & 8,953  & TO   & TO  & TO \\
    &&&& 20,895 & 9,211  & \b{7,099}  & TO   & TO  & TO \\\addlinespace[0.5em]
  \bottomrule
\end{tabular}
		\label{tab:results_cps_all}
	\end{table}
\end{landscape}

\begin{landscape}
	\begin{table}
		\centering
		\small
		\caption{Benchmark results for case studies from the quantitative verification benchmark set (part 1).}
		\begin{threeparttable}

\begin{tabular}{@{}lrrr rrrrrrr@{}} \toprule
  & \multicolumn{3}{c}{Comparision}
&\multicolumn{2}{c}{Single Feature} & \multicolumn{3}{c}{Linear} & \multicolumn{2}{c}{Quadratic}\\
\cmidrule(r){2-4} \cmidrule(r){5-6} \cmidrule(r){7-9} \cmidrule(r){10-11}
Case Study & States & BDD & MinSize
& Ax.Al. & Categ. & LinSVM & LogReg & OC1   & Poly & PolyPrio1\\ \midrule
\multirow{2}{*}[0pt]{triangle-tireworld.9} & \multirow{2}{*}[0pt]{48}& \multirow{2}{*}[0pt]{51}& \multirow{2}{*}[0pt]{\b{17}}
   &      27&      28&      25&      23&      19&      25&      \b{17}\\
&&&&      31&      31&      21&      25&      23&      \b{17}&      \b{17}\\\addlinespace[0.5em]
\multirow{2}{*}[0pt]{pacman.5} & \multirow{2}{*}[0pt]{232}& \multirow{2}{*}[0pt]{330}& \multirow{2}{*}[0pt]{\b{37}}
   &      53&      43&      51&      49&      55&      47&      \b{37}\\
&&&&      81&      81&      71&      59&      81&      \b{37}&      \b{37}\\\addlinespace[0.5em]
\multirow{2}{*}[0pt]{rectangle-tireworld.11} & \multirow{2}{*}[0pt]{241}& \multirow{2}{*}[0pt]{495}& \multirow{2}{*}[0pt]{481}
   &     481&  \b{373}\tnote{1}
                     &     481&     481&     481&     481&     481\\
&&&&     481&     481&     481&     481&     481&     481&     481\\\addlinespace[0.5em]
\multirow{2}{*}[0pt]{philosophers-mdp.3} & \multirow{2}{*}[0pt]{344}& \multirow{2}{*}[0pt]{295}& \multirow{2}{*}[0pt]{59}
   &     391&     181&     381&     377&     333&     315&     251\\
&&&&     403&     307&     375&     367&     393&     251&     \b{223}\\\addlinespace[0.5em]
\multirow{2}{*}[0pt]{firewire\_abst.3.rounds} & \multirow{2}{*}[0pt]{610}& \multirow{2}{*}[0pt]{295}& \multirow{2}{*}[0pt]{\b{25}}
   &      \b{25}&      \b{25}&      \b{25}&      \b{25}&      \b{25}&      \b{25}&      \b{25}\\
&&&&      \b{25}&      \b{25}&      \b{25}&      \b{25}&      \b{25}&      \b{25}&      \b{25}\\\addlinespace[0.5em]
\multirow{2}{*}[0pt]{rabin.3} & \multirow{2}{*}[0pt]{704}& \multirow{2}{*}[0pt]{303}& \multirow{2}{*}[0pt]{\b{23}}
   &     111&     187&      51&      43&      29&      69&      27\\
&&&&     175&     137&      31&      29&      45&      \b{23}&      \b{23}\\\addlinespace[0.5em]
\multirow{2}{*}[0pt]{ij.10} & \multirow{2}{*}[0pt]{1,013}& \multirow{2}{*}[0pt]{436}& \multirow{2}{*}[0pt]{19}
&   1,291&   1,291&     907&     753&     771&     897&     209\\
&&&&   1,405&   1,405&     893&     735&   1,131&     \b{141}&     177\\\addlinespace[0.5em]
\multirow{2}{*}[0pt]{\scriptsize zeroconf.1000.4.true.correct\_max} & \multirow{2}{*}[0pt]{1,068}& \multirow{2}{*}[0pt]{535}& \multirow{2}{*}[0pt]{\b{45}}
&      83&      83&      67&      63&      49&      75&      57\\
&&&&      79&      79&      47&      \b{45}&      71&      \b{45}&      \b{45}\\\addlinespace[0.5em]
\multirow{2}{*}[0pt]{blocksworld.5} & \multirow{2}{*}[0pt]{1,124}& \multirow{2}{*}[0pt]{3,985}& \multirow{2}{*}[0pt]{367}
&   1,687&   1,308&   1,515&   1,407&   1,583&   1,451&     891\\
&&&&   1,771&   1,649&   1,535&   1,405&   1,719&     521&     \b{513}\\\addlinespace[0.5em]
\bottomrule
\end{tabular}

\begin{tablenotes}
\item[1] this is better than the minimum size as it uses non-binary splits
\end{tablenotes}
\end{threeparttable}
		\label{tab:results_storm_all1}
	\end{table}
\end{landscape}

\begin{landscape}
	\begin{table}
		\centering
		\small
		\caption{Benchmark results for case studies from the quantitative verification benchmark set (part 2).}
		\begin{tabular}{@{}lrrr rrrrrrr@{}} \toprule
  & \multicolumn{3}{c}{Comparision}
  &\multicolumn{2}{c}{Single Feature} & \multicolumn{3}{c}{Linear} & \multicolumn{2}{c}{Quadratic}\\
 \cmidrule(r){2-4} \cmidrule(r){5-6} \cmidrule(r){7-9} \cmidrule(r){10-11}
  Case Study & States & BDD & MinSize
    & Ax.Al. & Categ. & LinSVM & LogReg & OC1   & Poly & PolyPrio1\\ \midrule
\multirow{2}{*}[0pt]{cdrive.10} & \multirow{2}{*}[0pt]{1,921}& \multirow{2}{*}[0pt]{5,134}& \multirow{2}{*}[0pt]{1,903}
   &   2,401&   3,122&   2,401&   2,401&\b{2,257}&   2,401&   2,401\\
&&&&   2,089&   2,037& TO & TO &   TO & TO & TO \\\addlinespace[0.5em]
\multirow{2}{*}[0pt]{consensus.2.disagree} & \multirow{2}{*}[0pt]{2,064}& \multirow{2}{*}[0pt]{138}& \multirow{2}{*}[0pt]{25}
   &      67&      67&      75&      69&      69&      57&      51\\
&&&&     105&     105&     105&      93&      95&      35&   \b{33}\\\addlinespace[0.5em]
\multirow{2}{*}[0pt]{beb.3-4.LineSeized} & \multirow{2}{*}[0pt]{4,275}& \multirow{2}{*}[0pt]{913}& \multirow{2}{*}[0pt]{\b{57}}
&      65&      70&      65&      65&      63&      65&      59\\
&&&&      85&      76&      \b{57}&      \b{57}&      89&      \b{57}&      \b{57}\\\addlinespace[0.5em]
\multirow{2}{*}[0pt]{csma.2-4.some\_before} & \multirow{2}{*}[0pt]{7,472}& \multirow{2}{*}[0pt]{1,059}& \multirow{2}{*}[0pt]{\b{65}}
   &     103&     103&     107&     105&      89&     103&      79\\
&&&&     185&     185&      85&  \b{65}&     177&  \b{65}&      \b{65}\\\addlinespace[0.5em]
\multirow{2}{*}[0pt]{eajs.2.100.5.ExpUtil} & \multirow{2}{*}[0pt]{12,627}& \multirow{2}{*}[0pt]{1,315}& \multirow{2}{*}[0pt]{65}
   &     167&     160&     173&     161&     135&     133&     141\\
&&&&     167&     167&     167&     157&     133&     133&     \b{125}\\\addlinespace[0.5em]
\multirow{2}{*}[0pt]{elevators.a-11-9} & \multirow{2}{*}[0pt]{14,742}& \multirow{2}{*}[0pt]{6,750}& \multirow{2}{*}[0pt]{129}
   &  16,341&  16,413&  11,243&   9,865&  13,619&9,779&2,859\\
&&&&  17,809&  17,495&  11,505&   9,955&  16,423 & 2,023 & \b{1,919}\\\addlinespace[0.5em]
\multirow{2}{*}[0pt]{\scriptsize exploding-blocksworld.5} & \multirow{2}{*}[0pt]{76,741}& \multirow{2}{*}[0pt]{3,447}& \multirow{2}{*}[0pt]{149}
&  16,913&   8,138&   4,503&   2,687&   5,993&   4,511&  \b{829}\\
&&&&  20,273&   8,571&   5,307&   2,845&   6,893&TO&TO\\\addlinespace[0.5em]
\multirow{2}{*}[0pt]{echoring.MaxOffline1} & \multirow{2}{*}[0pt]{104,892}& \multirow{2}{*}[0pt]{43,165}& \multirow{2}{*}[0pt]{801}
&   2,101&   2,251&   1,629&   1,625&   1,627&2,005&\b{1,431}\\
&&&&   5,031&TO&TO&TO&TO&TO&TO\\\addlinespace[0.5em]
\multirow{2}{*}[0pt]{wlan\_dl.0.80.deadline} & \multirow{2}{*}[0pt]{189,641}& \multirow{2}{*}[0pt]{1,541}& \multirow{2}{*}[0pt]{175}
&   3,369&   3,369&   2,821&   2,563&     701&     693&     667\\
&&&&   3,675&   3,675&   2,841& 2,621 & 1,049 & \b{523} & TO\\\addlinespace[0.5em]
\multirow{2}{*}[0pt]{pnueli-zuck.5} & \multirow{2}{*}[0pt]{303,427}& \multirow{2}{*}[0pt]{\b{50,128}}& \multirow{2}{*}[0pt]{173}
& 171,371& 171,371& 156,165&150,341&125,421& 114,979&  83,219\\
&&&& 263,955& 263,955& 221,645& 214,801 & 221,645 &  95,879 & 83,951\\\addlinespace[0.5em]
  \bottomrule
\end{tabular}
		\label{tab:results_storm_all2}
	\end{table}
\end{landscape}

}
{}
\end{document}